\documentclass[10pt,twocolumn,letterpaper]{article}
\ifdefined\pdfsuppressptexinfo\pdfsuppressptexinfo=-1 \fi
\PassOptionsToPackage{table}{xcolor}
\usepackage{cvpr}              
\usepackage[T1]{fontenc}
\usepackage{amsmath}
\usepackage{amsfonts}
\usepackage{amssymb}
\usepackage{xspace}
\usepackage{url}

\makeatletter
\DeclareRobustCommand\onedot{\futurelet\@let@token\@onedot}
\def\@onedot{\ifx\@let@token.\else.\null\fi\xspace}
\makeatother

\usepackage{graphicx}
\usepackage{booktabs}

\usepackage[table,dvipsnames]{xcolor}
\usepackage{colortbl}

\usepackage[accsupp]{axessibility}

\definecolor{cvprblue}{rgb}{0.21,0.49,0.74}
\usepackage[breaklinks,colorlinks,allcolors=cvprblue]{hyperref}

\newcommand{\ourmethod}{\textup{\textsc{EvDualDiff}}\xspace}
\newcommand{\ourdataset}{EvDual\xspace}
   \newcommand{\ourencoder}{Time-to-Brightness Encoder\xspace}
\newcommand{\PAR}[1]{\noindent{\bf #1}}

\usepackage{pifont}   \usepackage{diagbox}
\usepackage{adjustbox}
\usepackage{multirow}
\usepackage{dsfont}

\definecolor{tableHeadGray}{gray}{.9}
\definecolor{tableSubHeadGray}{gray}{0.95}
\newcommand{\pub}[1]{\kern1pt{\color{gray}{\tiny{[{#1}]\!}}}}
\usepackage{arydshln} \definecolor{lightgreen}{RGB}{220, 250, 220}
\definecolor{lightgreen2}{RGB}{235, 245, 230}

\colorlet{colorFst}{Green!25}       \colorlet{colorSnd}{SpringGreen!45} \colorlet{colorTrd}{Yellow!30}      \colorlet{colorLow}{darkgray!30}
\newcommand{\fs}{\cellcolor{colorFst}\bf}   \newcommand{\nd}{\cellcolor{colorSnd}}      
      \newcommand{\thickhline}{\noalign{\hrule height 0.9pt}}
\newcolumntype{?}{!{\vrule width 0.9pt}}

\makeatletter
\g@addto@macro\normalsize{  \setlength\abovedisplayskip{5pt plus 1pt minus 2pt}\setlength\belowdisplayskip{5pt plus 1pt minus 2pt}  \setlength\abovedisplayshortskip{2pt plus 1pt}\setlength\belowdisplayshortskip{4pt plus 1pt minus 2pt}}
\def\cvprsection{\@startsection {section}{1}{\z@}{-6pt plus -2pt minus -1pt}{4pt}{\large\bf}}
\def\cvprsubsection{\@startsection {subsection}{2}{\z@}{-5pt plus -2pt minus -1pt}{3pt}{\elvbf}}
\makeatother

\def\confName{CVPR}
\def\confYear{2027}

\title{Dual-Stream Event-to-Video Reconstruction}

\author{Lei Sun$^{1,}$\thanks{Equal contribution.}\quad
Zhendong Li$^{1,*}$\quad
Chenxin Shao$^{2}$\quad
Weilun Li$^{2}$\\
Ruibo Ming$^{1}$\quad
Kaiwei Wang$^{2}$\quad
Jinjin Gu$^{1}$\quad
Danda Pani Paudel$^{1}$\quad
Luc Van Gool$^{1}$\\[3pt]
{ $^{1}$\,INSAIT, Sofia University ``St. Kliment Ohridski''}\\
{ $^{2}$\,Zhejiang University}
}

\begin{document}
\maketitle

\begin{abstract}

Event cameras capture brightness changes with high temporal resolution and dynamic range, but converting event streams into videos remains fundamentally ill-posed due to the lack of absolute intensity information.
Resolving this ambiguity calls for a brightness reference that complements the intensity changes recorded by events.
Inspired by the complementary spatial and temporal processing of the parvocellular and magnocellular pathways in human vision, we propose \ourmethod, a dual-stream framework that combines a brightness reference from temporal-mapping events with motion-event cues for video reconstruction.
Conditioned on both streams, our video diffusion model autoregressively reconstructs long sequences from a single temporal map. Photometric constraints and finetuning on self-generated rollouts reduce accumulated drift.
At the hardware level, our approach requires only a minimal light-modulation mechanism on the lens of an event camera. We build this setup and collect a dataset comprising temporal-mapping events, motion events and synchronised ground-truth RGB frames. Experiments on a public benchmark and on our collected \ourdataset dataset demonstrate that \ourmethod outperforms existing event-to-video reconstruction methods in fidelity, perceptual quality and video-level realism. The code and dataset will be released.
\end{abstract}

\section{Introduction}
\label{sec:intro}
Drawing upon the biological mechanisms of human sight, the silicon retina~\cite{mahowald1994silicon} established the conceptual framework for neuromorphic sensors, eventually evolving into what are now widely recognised as event cameras, or Dynamic Vision Sensors (DVS). DVS records \textit{relative} brightness changes as event streams with high temporal resolution (in the order of $\mu$s) and a wide dynamic range (over $120$~dB). DAVIS sensors provide both events and intensity frames, but representative designs have larger pixels and lower resolution than modern event-only DVS~\cite{gallego2020event}, which can limit their use in applications requiring high spatial resolution. Here, we study how high-resolution, event-only DVS can acquire a brightness reference without native frame readout or changes to their pixel circuitry. Among the applications of event cameras, event-to-video reconstruction plays a central role, as recovering intensity frames enables compatibility with conventional vision pipelines.

Learning-based methods, from E2VID~\cite{rebecq2019events,rebecq2019high} and its successors~\cite{ercan2024hypere2vid,cadena2023sparse,Cadena2021Spade-E2VID} to recent diffusion-based ones~\cite{Li2025EvDiffEV,Liang2024E2VIDiffPE,unie2f}, reconstruct video from events alone. This remains ill-posed because the intensity \textit{itself} is not measured, and the resulting ambiguity in initial brightness can lead to intensity drift and ghosting artefacts.

\begin{figure}[t]
    \centering
    \includegraphics[width=\linewidth]{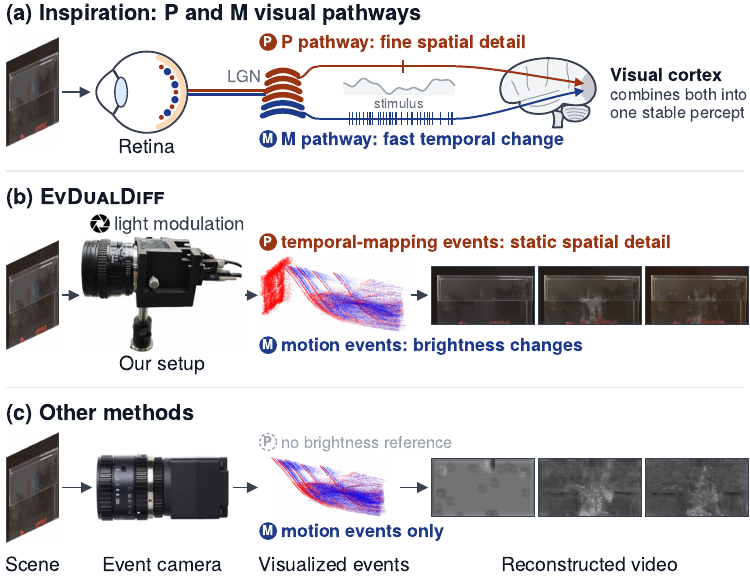}
\caption{\textbf{Dual-stream event-to-video reconstruction.} (a)~The P and M visual pathways carry fine spatial detail and fast temporal change. (b)~Likewise, \ourmethod\ records temporal-mapping events and motion events with one sensor and reconstructs video from both. (c)~Previous methods~\cite{E2VID+} that see only motion events lose the static background and the brightness level.}
    \label{fig:teaser}
\end{figure}

The human visual system processes spatial and temporal information through complementary pathways. This parallel organisation is exemplified by the magnocellular (M) and parvocellular (P) pathways~\cite{livingstone1988segregation}, as sketched in Fig.~\ref{fig:teaser}(a). The M pathway is particularly sensitive to rapid temporal changes, while the P pathway supports fine spatial processing. This functional complementarity motivates combining spatial appearance with temporal changes for video reconstruction.

Guided by this principle, we propose \ourmethod, a dual-stream framework that combines temporal-mapping events~\cite{bao2024temporal} and motion events from a single sensor for video reconstruction. As Fig.~\ref{fig:teaser}(b) shows, a light-modulation device in the lens raises the transmittance from $0\%$ to $100\%$ within a few milliseconds and then keeps it constant. The sweep generates temporal-mapping events that provide a brightness reference, while subsequent motion events capture scene dynamics. In contrast, E2VID+~\cite{E2VID+} uses only motion events and produces a nearly blank initial frame and low-contrast reconstructions in Fig.~\ref{fig:teaser}(c). We condition a video diffusion model on these complementary brightness and motion cues for reconstruction.
To effectively encode the temporal-mapping events, we further design the \ourencoder, a temporal-mapping-adapted VAE encoder that learns a compact latent representation aligned with the generative prior.

To reconstruct long, temporally dense videos beyond the $13$-frame chunks our model processes, we propose an autoregressive reconstruction strategy. We initialise reconstruction with a single measured temporal map and condition each subsequent chunk on motion events and a self-made map generated from the last reconstructed frame through the degradation model. Reusing generated references introduces photometric and content drift. We introduce low-band freezing and colour locking to stabilise brightness, and finetune the control branch on self-generated rollouts to reduce content drift from imperfect references. Our evaluated rollouts reach up to $43$ chunks and $517$ distinct frames from one initial temporal map.

Additionally, we build a shared-lens beam-splitter system and collect \ourdataset, pairing temporal-mapping and motion events from one acquisition with synchronised RGB ground-truth videos of dynamic real-world scenes.

In summary, the main contributions of this work are:
\begin{itemize}
    \item Inspired by the complementary processing of the M and P pathways, we formulate video reconstruction with complementary brightness and motion conditions. \ourmethod\ integrates both event streams into a video diffusion prior through the \ourencoder\ and a control branch.
    \item We enable autoregressive reconstruction from a single temporal map. Two inference-time photometric constraints and finetuning on self-generated rollouts reduce brightness and content drift across video chunks.
    \item We build a shared-lens beam-splitter capture system and collect \ourdataset\ with synchronised temporal-mapping events, motion events and ground-truth RGB videos. Experiments on \ourdataset\ and the public BS-ERGB benchmark~\cite{tulyakov2022time} demonstrate improved reconstruction fidelity and perceptual quality.
\end{itemize}

\section{Related Work}
\label{sec:related_work}

\subsection{Event-to-Video Reconstruction}
High temporal resolution, low redundancy, and wide dynamic range make event cameras useful for SLAM~\cite{vidal2018ultimate}, feature tracking~\cite{messikommer2023data}, motion estimation~\cite{sun2026event}, deblurring~\cite{vitoria2022event,sun2022event,sun2025ntire}, video interpolation~\cite{tulyakov2022time,sun2024unified,chen2025repurposing}, optical flow~\cite{ye2025towards,gehrig2021raft}, autofocus~\cite{lin2022autofocus,bao2023improving}, and segmentation~\cite{zhang2023multi}. Event-to-video reconstruction recovers intensity frames for conventional vision pipelines~\cite{rebecq2019high,bao2024temporal}. Early approaches used physical models with filtering or integration~\cite{bardow2016simultaneous,munda2018real}. E2VID~\cite{rebecq2019events} introduced recurrent U-Net regression~\cite{ronneberger2015u}, FireNet~\cite{2020FireNet} reduced the model size, and E2VID+ and FireNet+~\cite{E2VID+} retrained both on improved synthetic data to narrow the sim-to-real gap. Such regression methods typically produce greyscale videos with limited perceptual realism. EvDiff~\cite{Li2025EvDiffEV}, E2VIDiff~\cite{Liang2024E2VIDiffPE}, UniE2F~\cite{unie2f}, and LongE2V~\cite{fan2026longe2v} instead use pretrained diffusion priors for video reconstruction. However, these methods rely on motion events alone or, as UniE2F also allows, on an RGB reference frame, which an event-only sensor does not provide; IE2Video~\cite{torbunov2025ie2video} and Engram-E2VID~\cite{ji2026engram} likewise start from an RGB frame.
AE2VID~\cite{bai2026ae2vid} also combines aperture-modulation events with motion events, but periodically refreshes its intensity references, interrupting unmodulated motion-event acquisition. Its bidirectional reconstruction additionally requires a future reference. Our framework instead uses one initial temporal map and subsequent motion events, avoiding repeated modulation and future brightness measurements. Since the repeated references required by AE2VID are unavailable in our single-map setting, its original pipeline cannot be evaluated directly under the same single-map input protocol.

\subsection{Autoregressive Video Diffusion Models}
Video diffusion models jointly denoise video frames to model spatial appearance and temporal dependencies~\cite{ho2022video,svd}.
For long video generation, MCVD~\cite{voleti2022mcvd} introduced an early blockwise autoregressive formulation that conditions each new block on preceding frames.
SEINE~\cite{chen2023seine} uses random-mask conditioning for video transitions and autoregressive prediction.
ViD-GPT~\cite{gao2024vid} introduces causal temporal attention and KV caching to retain past context while reducing redundant computation.
StreamingT2V~\cite{henschel2025streamingt2v} combines short-term attention with long-term appearance memory to maintain continuity across generated chunks.
StarGen~\cite{zhai2025stargen} conditions successive chunks on temporally overlapping frames and spatially adjacent views warped in 3D for consistent scene generation.

In event-based video reconstruction, EvDiff~\cite{Li2025EvDiffEV} uses a recurrent event encoder with one-step diffusion for sequential reconstruction.
E2VIDiff~\cite{Liang2024E2VIDiffPE} reconstructs fixed-length chunks, while UniE2F~\cite{unie2f} also demonstrates sliding-window extension to $23$ frames using its prediction mode.

\section{Methodology}
\label{sec:method}

\subsection{Preliminaries}
\label{sec:pre}
An event stream $\mathcal{E}=\{e_i\}_{i=1}^{N}$ records asynchronous brightness changes, where $e_i=(x_i,y_i,p_i,t_i)$ specifies pixel coordinates $(x_i,y_i)$, polarity $p_i\in\{-1,+1\}$, and timestamp $t_i$.
An event is triggered when the change in log intensity exceeds a contrast threshold $C$:
\begin{equation}
\left| 
\log \mathcal{I}(x_i, y_i, t_i)
- 
\log \mathcal{I}(x_i, y_i, t_i - \delta t)
\right| 
\ge C,
\label{eq:event_trigger}
\end{equation}
where $\mathcal{I}(x,y,t)$ denotes latent image intensity.
The interval $\delta t$ is measured from the previous event at the same pixel.
Event-to-video reconstruction maps $\mathcal{E}$ to $J$ intensity frames, $\{\mathbf{I}_j\}_{j=1}^{J}=\Phi(\mathcal{E})$.
Without an intensity reference, reconstruction remains ambiguous and can produce artefacts, as illustrated in Fig.~\ref{fig:teaser}(c).

EvTemMap~\cite{bao2024temporal} varies optical transmittance through controlled light modulation to generate temporal-mapping events.
Under the calibrated modulation, brighter pixels trigger their first positive event earlier, so the time $t^{+}(x,y)$ of this event estimates the brightness of the pixel:
\begin{equation}
\mathbf{I}_{tm}(x,y) = \psi\!\left(t^{+}(x,y)\right),
\label{eq:temporal_mapping}
\end{equation}
where $\psi(\cdot)$ is a hardware-dependent calibration mapping determined by the modulation pattern, photoelectric conversion efficiency, and pixel circuitry.
In the idealised case, a pixel fires its first positive event when the product of its intensity and the elapsed time reaches a threshold $\vartheta$, so $\psi(t^{+})=\vartheta/t^{+}$.
This mapping provides a brightness reference, although physical and electronic imperfections make it noisy in practice.

\begin{figure*}[t]
  \centering
  \includegraphics[width=\textwidth]{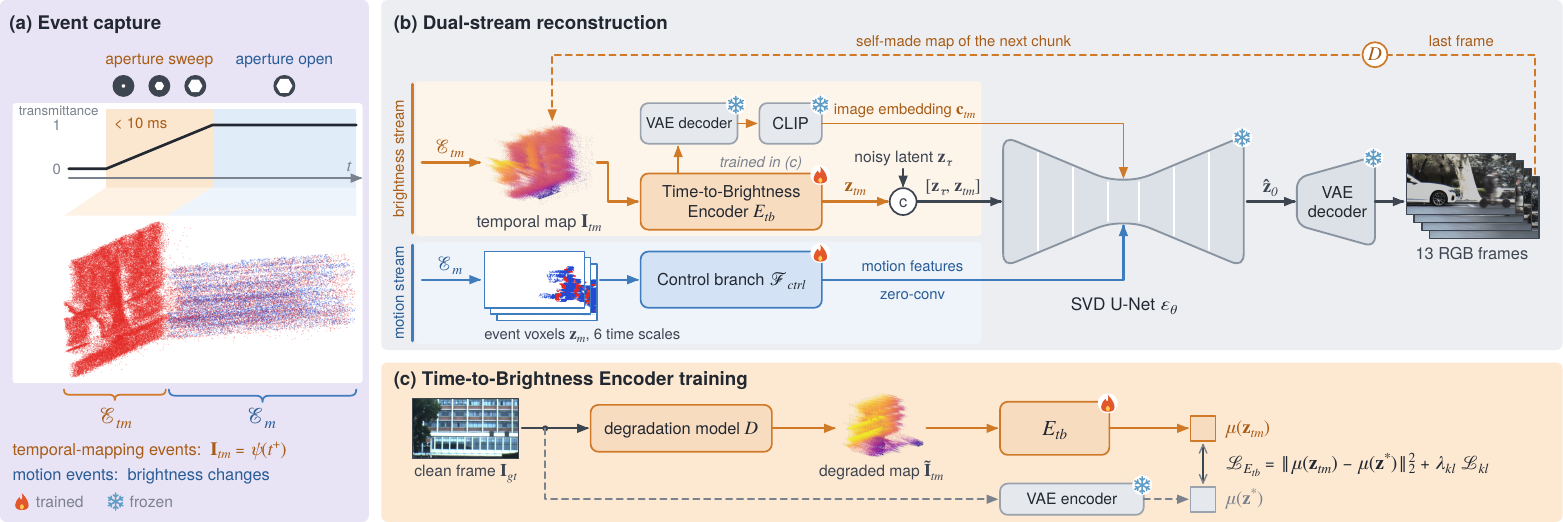}
\caption{\textbf{Overview of \ourmethod.} (a)~Event capture: the aperture sweep yields the temporal-mapping events $\mathcal{E}_{tm}$, the open aperture the motion events $\mathcal{E}_m$. (b)~The Time-to-Brightness Encoder $E_{tb}$ encodes the temporal map and CLIP embeds its decoded latent, the control branch $\mathcal{F}_{ctrl}$ injects $\mathcal{E}_m$ into the frozen SVD U-Net, and the degradation model $D$ turns the last frame into the self-made map of the next chunk (dashed). (c)~$E_{tb}$ is trained to map frames degraded by $D$ to the latent of the clean frame. The brightness stream is drawn in orange and the motion stream in blue.}
  \label{fig:pipeline}
\end{figure*}

\subsection{\ourmethod Pipeline}
\label{sec:overview}

In our dual-stream pipeline, we complement the motion events $\mathcal{E}_m$ of Eq.~\ref{eq:event_trigger} with temporal-mapping events $\mathcal{E}_{tm}$, which provide a temporal map through Eq.~\ref{eq:temporal_mapping}, and condition the reconstruction $\Phi$ on both streams:
\begin{equation}
\{\mathbf{I}_j\}_{j=1}^{J}
= \Phi_{\theta}\!\left(\mathcal{E}_m \mid \mathbf{I}_{tm}\right),
\label{eq:dual_stream_reconstruction}
\end{equation}
where $\mathbf{I}_{tm}$ is the noisy temporal map and $\theta$ denotes the reconstruction model parameters.
Direct polarity integration remains unreliable because noise and trailing events accumulate.
A diffusion prior suppresses these artefacts and promotes spatiotemporal consistency.

We implement this formulation with the widely adopted Stable Video Diffusion (SVD)~\cite{svd} and a ControlNet-style control branch $\mathcal{F}_{ctrl}$~\cite{zhang2023adding}.

\PAR{Event Capture.}~Fig.~\ref{fig:pipeline}(a) shows the capture process: a motorised aperture sweeps the lens transmittance in less than $10\,\mathrm{ms}$, which yields the temporal-mapping events, and then stays fully open while the same sensor records motion events.
As the sweep is this short, we treat the scene as static during it, as for a frame with the same exposure time.
The aperture is synchronised with the DVS, so temporal-mapping events $\mathcal{E}_{tm}$ and motion events $\mathcal{E}_m$ are separated by their acquisition intervals. During the short transition between the two, the event types overlap, and the mixed events are discarded. The recovered temporal map is therefore imperfect, and the Time-to-Brightness Encoder of Sec.~\ref{sec:degradation} is trained to tolerate such degradations.

\PAR{Event Encoding and Denoising.}~Both streams condition latent denoising through
\begin{equation}
\begin{aligned}
\mathbf{z}_{tm} &= E_{tb}(\mathbf{I}_{tm}), \quad
\mathbf{z}_m = E_{mot}(\mathcal{E}_m), \\
\hat{\boldsymbol{\epsilon}}_\tau
&= \epsilon_{\theta}\!\left([\mathbf{z}_\tau,\mathbf{z}_{tm}],\tau \mid \mathbf{z}_m,\mathbf{c}_{tm}\right),
\end{aligned}
\label{eq:dual_stream_denoising}
\end{equation}
where $\epsilon_{\theta}$ is the denoising network, $\mathbf{z}_\tau$ is the noisy video latent at diffusion timestep $\tau$ and $[\cdot,\cdot]$ denotes channel-wise concatenation, with $\mathbf{z}_{tm}$ repeated for every frame. The Time-to-Brightness Encoder $E_{tb}$ supplies the brightness condition, $\mathbf{c}_{tm}$ is the CLIP~\cite{radford2021learning} image embedding of the temporal map, which takes the place of the image conditioning of the backbone, and $E_{mot}$ stacks the motion events at six time scales~\cite{lu2024hr} and passes the resulting tensor $\mathbf{z}_m$ to the control branch $\mathcal{F}_{ctrl}$. The temporal map has a single channel: it is replicated to three channels for $E_{tb}$, and $\mathbf{c}_{tm}$ is computed on the image that the frozen VAE decoder renders from the latent of the map. Neither stream measures colour, so the chromaticity of the output is inferred by the model, which is trained against RGB frames. Starting from Gaussian noise, iterative denoising yields the video latent.

\subsection{Time-to-Brightness Encoder}
\label{sec:degradation}
The original VAE encoder~\cite{kingma2013auto,svd} is trained on clean natural images, whereas temporal-mapping images contain temporal- and intensity-domain degradations.
We construct a degradation model $D$ for temporal-mapping events to synthesise these effects and train a Time-to-Brightness Encoder adapted to the resulting maps (Fig.~\ref{fig:pipeline}(c)).

Starting from the luminance of high-quality images, we apply the diffraction blur, invert Eq.~\ref{eq:temporal_mapping}, perturb the temporal map with threshold variations, map it back to intensity, and add defective pixels and sensor noise.
This process builds on EvTemMap~\cite{bao2024temporal} and adds the diffraction blur of our modulated aperture, absent from its setup.

\PAR{Diffraction-Induced Blur During Modulation.}~The small aperture diameter $d(t)$ early in the sweep causes strong diffraction.
The Airy-disk radius is $\rho(t)\approx1.22\lambda f/d(t)$~\cite{smith2008modern}, with wavelength $\lambda$ and focal length $f$.
This wide point-spread function acts as a low-pass filter, and blur accumulated early in the sweep remains in the recovered intensity.
The blur of a pixel thus depends on when it fires, and hence on its brightness. We approximate it by a spatially uniform blur, a fixed kernel in $D$ plus Gaussian blur of random strength in encoder training.

We train $E_{tb}$ by aligning degraded-image latents with clean-image latents, avoiding a decoder round-trip:
\begin{equation}
\mathcal{L}_{E_{tb}} = \left\| \mu(\mathbf{z}_{tm}) - \mu(\mathbf{z}^{*}) \right\|^2_2 + \lambda_{kl} \mathcal{L}_{kl}.
\label{eq:tb_alignment}
\end{equation}
Here, $\mathbf{z}_{tm}=E_{tb}(\tilde{\mathbf{I}}_{tm})$, $\mathbf{z}^{*}=\mathrm{VAE}_{enc}(\mathbf{I}_{gt})$ is given by the frozen original encoder $\mathrm{VAE}_{enc}$ for the clean RGB target $\mathbf{I}_{gt}$, and $\mu$ selects the posterior mode.
$\mathcal{L}_{kl}$ is the KL divergence between Gaussians fitted to the channel statistics (mean and covariance over spatial positions) of the two latents, with $\lambda_{kl}=0.1$.
We supervise the mode because both control-branch training and inference consume it.
The input $\tilde{\mathbf{I}}_{tm}$ includes the conditioning noise augmentation that SVD applies before encoding, in control-branch training and at inference, whereas the target $\mathbf{z}^{*}$ is computed without it.
The applied noise level is still passed to the backbone as in SVD (see the supplementary material).
Degradation and optimisation details are in the supplementary material.

\begin{figure*}[t]
\centering
\includegraphics[width=\textwidth]{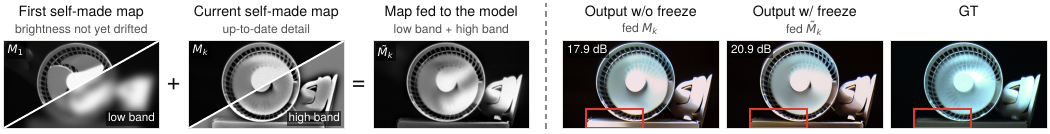}
\caption{\textbf{Low-band freezing for autoregressive conditioning.} Left: Eq.~\ref{eq:freeze} for $k=3$: the first self-made map $M_1$ supplies the low band and the current map $M_k$ the high band of the fed map $\tilde{M}_k$. Right: a frame reconstructed from $M_k$ and from $\tilde{M}_k$, with luminance PSNR, and the ground truth; the box marks the largest reduction of the luminance error. Maps are shown without sensor noise, the high band with enhanced contrast.}
\label{fig:freqsplit}
\end{figure*}

\begin{figure}[t]
  \centering
  \includegraphics[width=\columnwidth]{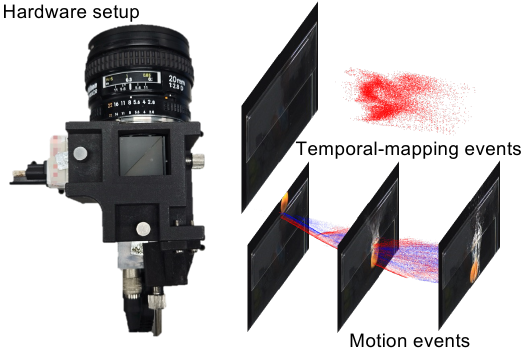}
\caption{\textbf{Capture system and the two event streams.} A motorised aperture and a 50R/50T beam splitter behind a shared lens give the event and RGB sensors the same optical path. The aperture sweep yields temporal-mapping events (top); once it is fully open, the sensor records motion events (bottom).}
  \label{fig:dataset}
\end{figure}

\subsection{Dual-Stream Video Reconstruction}
\label{sec:diffusion}
\label{sec:autoreg}
\label{sec:antidrift}

\PAR{Motion Events as Control.}~Event-image paired datasets are several orders of magnitude smaller than large-scale text-image datasets such as LAION-5B~\cite{schuhmann2022laion}.
To preserve the pretrained video prior, we freeze the diffusion backbone and train the control branch $\mathcal{F}_{ctrl}$ on paired event-image data~\cite{zhang2023adding}.
Zero-convolution layers inject multi-scale motion features into the frozen backbone, implementing the motion conditioning in Eq.~\ref{eq:dual_stream_denoising}.
Given a ground-truth video latent $\mathbf{z}_0$, Gaussian noise $\boldsymbol{\epsilon}\sim\mathcal{N}(0,\mathbf{I})$ and a noise level $\sigma$, we optimise the control branch with the denoising objective of the SVD backbone, which follows EDM~\cite{karras2022elucidating}:
\begin{equation}
\begin{aligned}
\mathcal{L} &= \mathbb{E}_{\mathbf{z}_{0}, \boldsymbol{\epsilon}, \sigma, \mathcal{E}_m} \left[ \tfrac{1+\sigma^2}{\sigma^2} \left\| \hat{\mathbf{z}}_0 - \mathbf{z}_0 \right\|^2_2 \right], \\
\hat{\mathbf{z}}_0 &= c_{skip}\,\mathbf{z}_\tau + c_{out}\,\hat{\boldsymbol{\epsilon}}_\tau,
\end{aligned}
\end{equation}
where $\mathbf{z}_\tau=\mathbf{z}_0+\sigma\boldsymbol{\epsilon}$ is the noisy latent, the timestep $\tau$ encodes $\sigma$, $\hat{\boldsymbol{\epsilon}}_\tau$ is the network output of Eq.~\ref{eq:dual_stream_denoising}, and $c_{skip}=1/(\sigma^2+1)$ and $c_{out}=-\sigma/\sqrt{\sigma^2+1}$ are the EDM preconditioning coefficients; the input scaling $c_{in}=1/\sqrt{\sigma^2+1}$ of $\mathbf{z}_\tau$ is absorbed into $\epsilon_{\theta}$.

\PAR{Autoregressive Video Reconstruction.}~We extend Eq.~\ref{eq:dual_stream_reconstruction} autoregressively by conditioning each video chunk on its motion events and a recursively updated map:
\begin{equation}
\begin{aligned}
\hat{\mathbf{V}}_k
&=\Phi_{\theta}\!\left(\mathcal{E}_m^{(k)}\mid M_k\right), \\
M_k
&=\begin{cases}
\mathbf{I}_{tm}, & k=0, \\
D\!\left(\hat{\mathbf{I}}_{k-1,\mathrm{last}}\right), & k\ge1.
\end{cases}
\end{aligned}
\label{eq:autoregressive_reconstruction}
\end{equation}
Here, $\hat{\mathbf{V}}_k$ denotes the reconstructed video chunk of $13$ frames and $\mathcal{E}_m^{(k)}$ the motion events measured between its first and last frame.
For subsequent chunks, the degradation model $D$ from Sec.~\ref{sec:degradation}, with all its noise components active as in training, converts the last frame of the previous chunk $\hat{\mathbf{I}}_{k-1,\mathrm{last}}$ into a \emph{self-made map} (dashed loop in Fig.~\ref{fig:pipeline}).
$D$ acts on the luminance of that frame, so every chunk is conditioned in the form that $E_{tb}$ and the control branch are trained on, a degraded single-channel map, rather than on a generated RGB frame, which neither sees in training (compared in the supplementary material).
Chunk~$k$ starts on that frame, so consecutive chunks share one frame and every later chunk adds $12$ new frames.
The shared frame is kept from the earlier chunk, so no frame is output or scored twice.
No further brightness measurement or ground-truth reference enters the rollout.

Reusing generated frames introduces (1) \emph{photometric drift} from accumulated brightness biases and (2) \emph{content drift} from rendering artefacts propagated through the self-made maps.
We address them with photometric constraints and finetuning on model rollouts, respectively.

\PAR{Photometric Stabilisation.}~Two constraints stabilise the self-made maps and reconstructed frames.
The \emph{low-band freeze} separates large-scale brightness from local detail, following frequency-split anchoring~\cite{li2026freqforcing}.
Let $L(M)=G_{\sigma_{\mathrm{lp}}}*M$ denote Gaussian low-pass filtering with scale $\sigma_{\mathrm{lp}}=32$~px.
The first self-made map $M_1$ is used unchanged.
For later chunks, we retain its low-frequency component and update the high-frequency component from the current self-made map:
\begin{equation}
  \tilde{M}_k=\mathrm{clamp}\big(L(M_1)+M_k-L(M_k),\,0,\,1\big),\qquad k\ge2.
  \label{eq:freeze}
\end{equation}
This preserves the large-scale brightness of the first self-made map while allowing edges and textures to evolve (Fig.~\ref{fig:freqsplit}).
The \emph{colour lock} aligns the statistics of each RGB channel of an output chunk with those of chunk~$0$:
$x\mapsto(x-\mu_{k,c})\,s_{0,c}/s_{k,c}+\mu_{0,c}$,
where $x$ is a channel-$c$ pixel value in chunk~$k$, and $\mu_{k,c}$ and $s_{k,c}$ are the corresponding mean and standard deviation.
On \ourdataset, chunk~$0$ is first scaled by a gain predicted from the luminance statistics of the temporal map, with coefficients fitted on the training split (see the supplementary material).
We apply this correction before exporting frames and constructing the next map, anchoring both the output and subsequent conditioning to the initial observation.
Neither of the two constraints changes the network weights.

\PAR{Anti-Drift Finetuning.}~Content drift stems from a conditioning mismatch: training maps come from clean frames, whereas inference maps come from generated frames.
Following the data-aggregation principle of DAgger~\cite{ross2011dagger}, and like Self Forcing~\cite{huang2025selfforcing}, which trains autoregressive video diffusion on its own rollouts, we construct self-made maps from rollouts generated by the base model over the training split.
Each self-made map replaces the clean-derived training map at the matching timestamp, while the measured motion events and clean reconstruction targets remain unchanged.
This exposes the model to conditioning errors encountered during autoregressive inference.
On \ourdataset, we also augment the maps with random exposure and blur perturbations to broaden the conditioning distribution, following conditioning augmentation in cascaded diffusion~\cite{ho2022cascaded}.
Finetuning retains the original diffusion objective and frozen backbone, adapting the control branch to reconstruct clean videos from imperfect maps.
A rollout schematic, the augmentation parameters and corpus details are in the supplementary material.

\PAR{Inference.}~At test time each frame is denoised at its native resolution on overlapping $512\times320$ views that are fused at every denoising step~\cite{bar2023multidiffusion} with Hann-window weights and a weight floor of $0.05$; the weights are detailed in the supplementary material and ablated in Sec.~\ref{sec:ablation}.

\section{Hardware Setup and \ourdataset Dataset}
\label{sec:dataset}

\begin{table*}[t]
\centering
\small
\setlength{\tabcolsep}{4pt}
\renewcommand\arraystretch{1.0}   \caption{\textbf{Quantitative comparison on BS-ERGB.} RGB and luminance metrics compare event-to-video reconstruction under the single-map protocol. CogVideoX-2B$^{*}$ denotes an event ControlNet trained on a frozen CogVideoX-2B and driven by events only. UniE2F$^\dagger$ denotes UniE2F retrained with the temporal map as its reference frame. Best and second-best results are highlighted.}
\begin{tabular}{r||ccccc?ccccc}
    \thickhline
    \rowcolor{tableHeadGray}
    & \multicolumn{5}{c?}{\textbf{RGB}} & \multicolumn{5}{c}{\textbf{Luminance ($Y$)}} \\
    \rowcolor{tableHeadGray}
    \multirow{-2}{*}{\textbf{Method}}
    & \textbf{PSNR $\uparrow$} & \textbf{SSIM $\uparrow$} & \textbf{LPIPS $\downarrow$} & \textbf{FID $\downarrow$} & \textbf{FVD $\downarrow$}
    & \textbf{PSNR $\uparrow$} & \textbf{SSIM $\uparrow$} & \textbf{LPIPS $\downarrow$} & \textbf{FID $\downarrow$} & \textbf{FVD $\downarrow$} \\
    \hline
    \rowcolor{tableSubHeadGray}
    SPADE-E2VID\pub{TIP2021}~\cite{Cadena2021Spade-E2VID} & 5.96 & 0.2924 & 0.7706 & 420.5 & 3736.2 & 6.01 & 0.3299 & 0.7256 & 369.7 & 2980.1 \\
    ET-Net\pub{ICCV2021}~\cite{ET-Net}                   & 9.52 & 0.2896 & 0.7157 & 408.1 & 3435.5 & 9.68 & 0.3274 & 0.6244 & 345.3 & 2713.9 \\
    \rowcolor{tableSubHeadGray}
    E2VID+\pub{ECCV2020}~\cite{E2VID+}                   & 10.55 & 0.2776 & 0.6718 & 408.6 & 3484.3 & 10.73 & 0.3103 & 0.5741 & 349.4 & 2700.2 \\
    FireNet\pub{WACV2020}~\cite{2020FireNet}             & 8.35 & 0.3053 & 0.7777 & 425.6 & 4051.7 & 8.45 & 0.3487 & 0.7316 & 382.7 & 3358.4 \\
    \rowcolor{tableSubHeadGray}
    EvDiff\pub{ECCV2026}~\cite{Li2025EvDiffEV}           & 12.12 & 0.2397 & \nd 0.5595 & 397.6 & 3525.1 & 12.31 & 0.2613 & 0.5126 & 365.3 & 3066.5 \\
    UniE2F\pub{arXiv2026}~\cite{unie2f}                  & 9.80 & 0.2901 & 0.8302 & 332.7 & 3017.5 & 10.16 & 0.3301 & 0.7311 & 312.7 & 2545.0 \\
    \rowcolor{tableSubHeadGray}
    LongE2V\pub{SIGGRAPH2026}~\cite{fan2026longe2v}      & 11.13 & 0.3267 & 0.7286 & 323.0 & 2491.4 & 11.36 & 0.3728 & 0.6229 & 285.2 & 1769.8 \\
    CogVideoX-2B$^{*}$\pub{ICLR2025}~\cite{yang2024cogvideox}          & 10.09 & 0.2458 & 0.6796 & 331.3 & 2929.7 & 10.24 & 0.2672 & 0.6040 & 329.4 & 2134.8 \\
    \hline\hline
    \multicolumn{11}{l}{\textcolor{gray}{\textit{Given the same initial temporal map as \ourmethod}}} \\
    \rowcolor{tableSubHeadGray}
    \quad EDI\pub{CVPR2019}~\cite{edi_pan}                             & 15.86 & 0.2764 & 0.7417 & 316.2 & 3441.1 & \nd 16.54 & 0.2988 & 0.6187 & \nd 247.7 & 2771.6 \\
    \quad UniE2F$^\dagger$\pub{arXiv2026}~\cite{unie2f}  & \nd 15.93 & \nd 0.4046 & 0.5693 & \nd 247.1 & \nd 1780.5 & 16.33 & \nd 0.4503 & \nd 0.5120 & 249.4 & \nd 1445.8 \\
    \hline\hline
    \textbf{\ourmethod~(Ours)}                           & \fs 16.86 & \fs 0.4489 & \fs 0.4192 & \fs 195.9 & \fs 1229.5 & \fs 17.35 & \fs 0.4973 & \fs 0.3646 & \fs 188.4 & \fs 953.6 \\
    \thickhline
\end{tabular}
\label{tab:bsergb}
\vspace{-1pt}   \end{table*}

\begin{figure*}[t]
\centering
\setlength{\tabcolsep}{0.9pt}
\renewcommand{\arraystretch}{0.1}

\setlength{\fboxsep}{0pt}\setlength{\fboxrule}{0.3pt}
\newcommand{\tile}[1]{\fcolorbox{black!30}{white}{#1}}
\newcommand{\lab}[1]{\multicolumn{1}{c}{\footnotesize #1}}

\begin{tabular}{cccccc}
\tile{\includegraphics[width=0.161\textwidth]{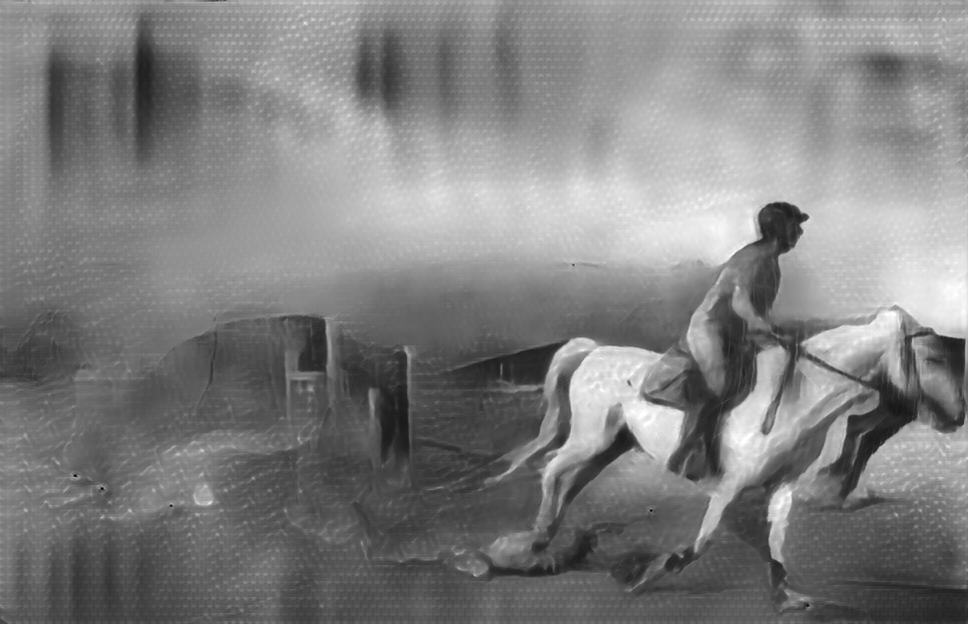}} & \tile{\includegraphics[width=0.161\textwidth]{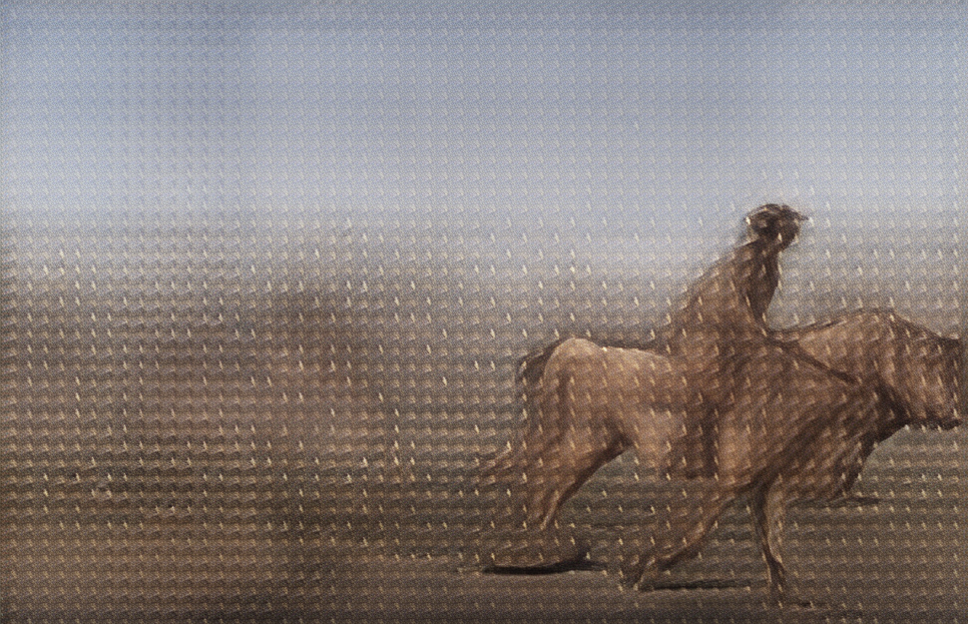}} & \tile{\includegraphics[width=0.161\textwidth]{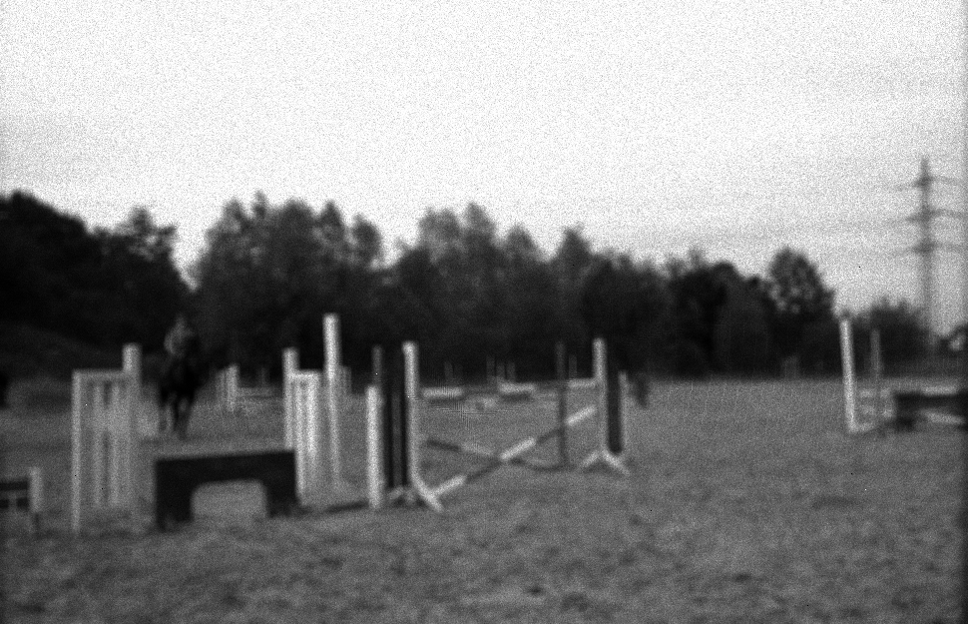}} & \tile{\includegraphics[width=0.161\textwidth]{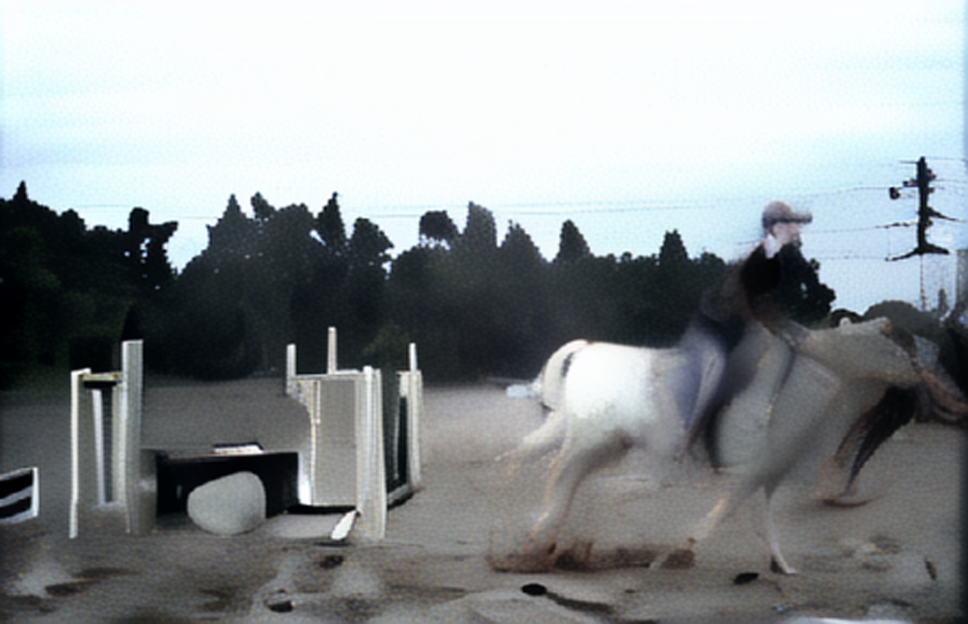}} & \tile{\includegraphics[width=0.161\textwidth]{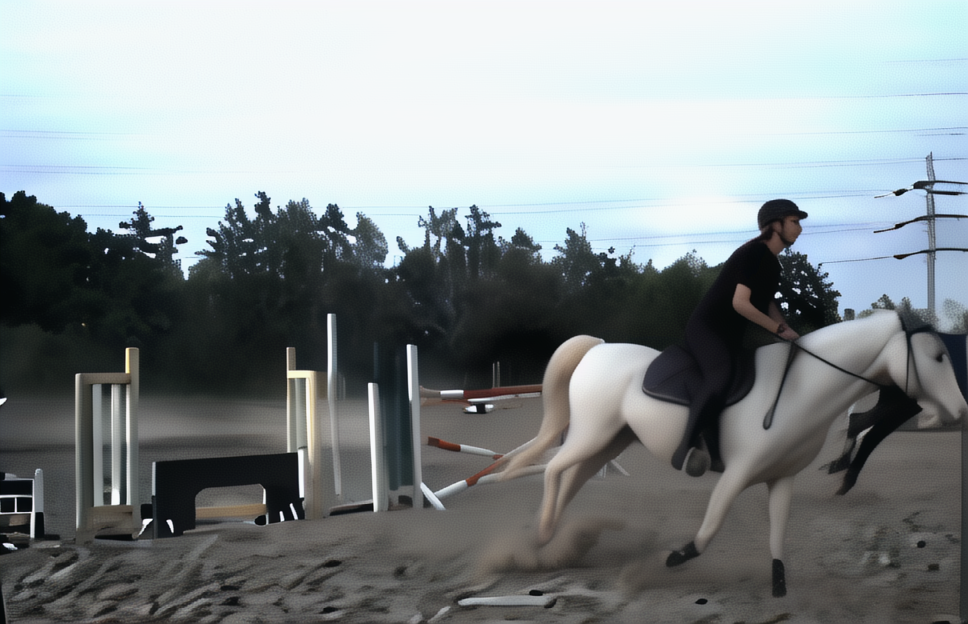}} & \tile{\includegraphics[width=0.161\textwidth]{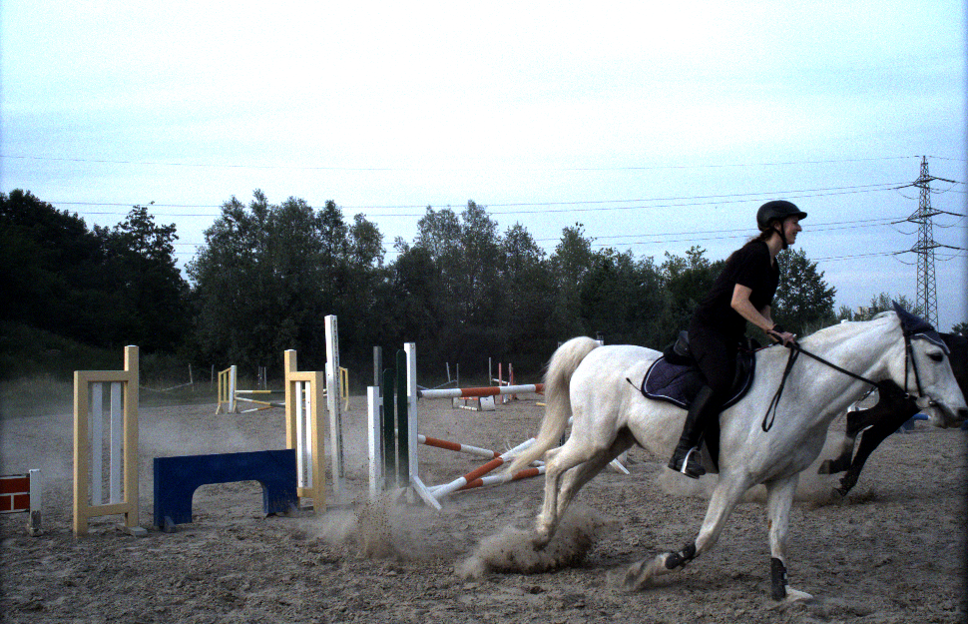}}\\[1.5pt]
\tile{\includegraphics[width=0.161\textwidth]{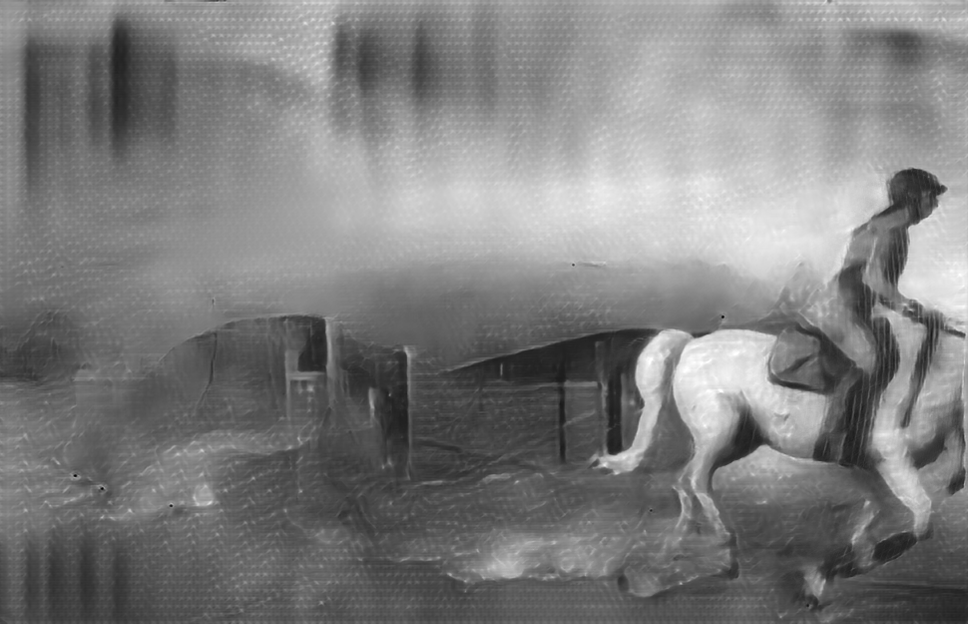}} & \tile{\includegraphics[width=0.161\textwidth]{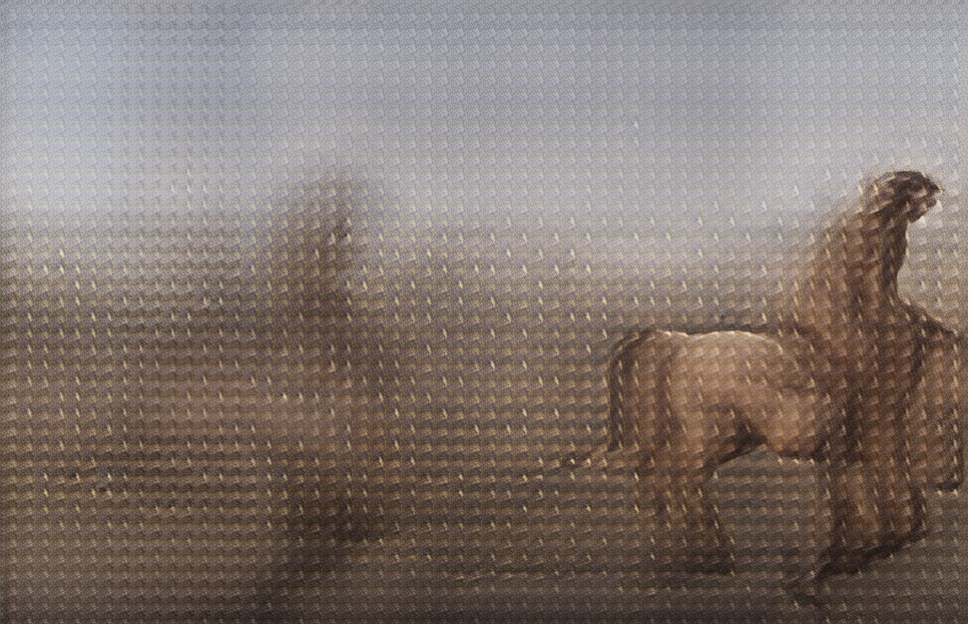}} & \tile{\includegraphics[width=0.161\textwidth]{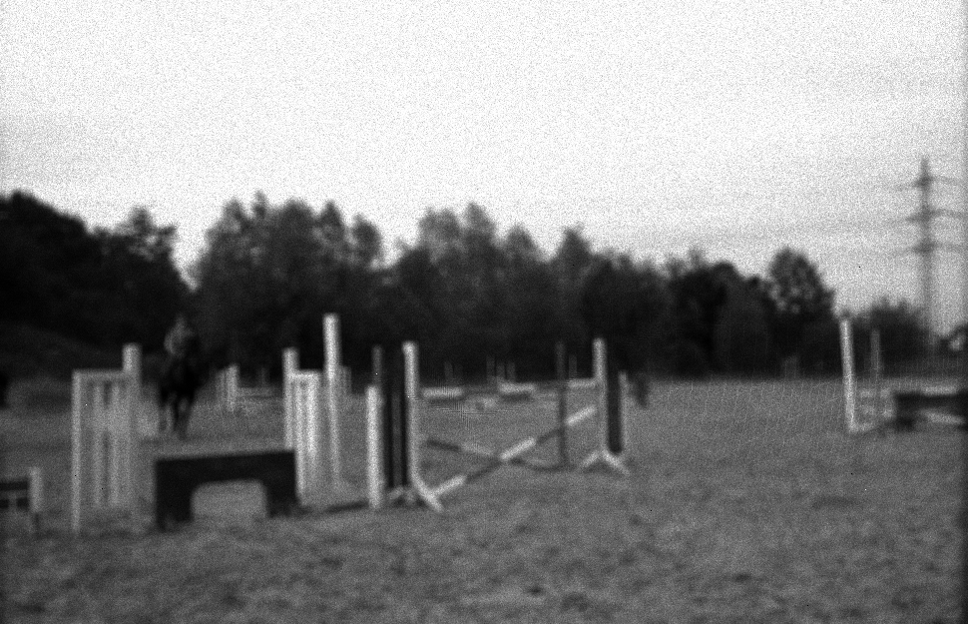}} & \tile{\includegraphics[width=0.161\textwidth]{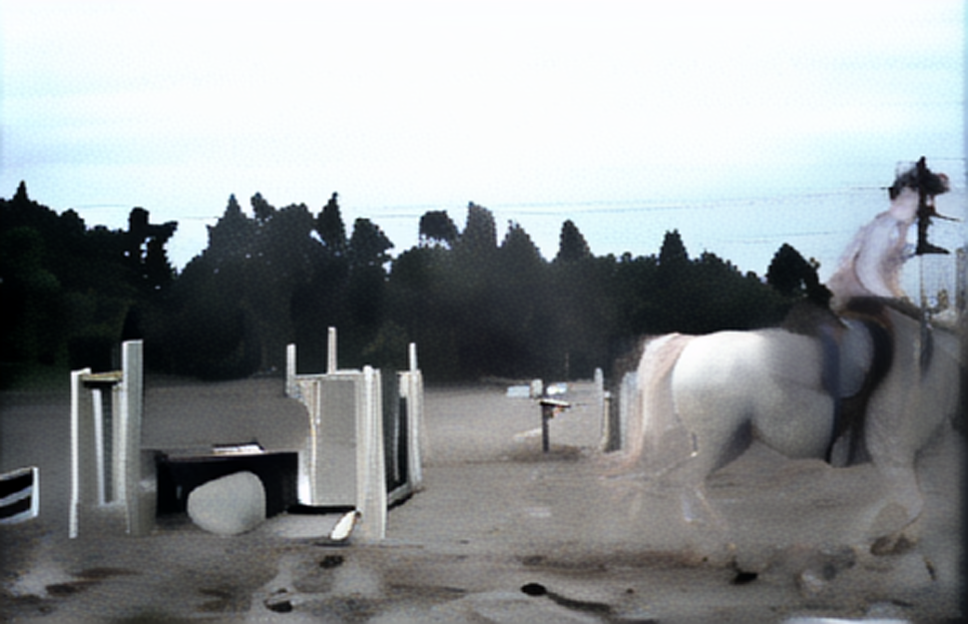}} & \tile{\includegraphics[width=0.161\textwidth]{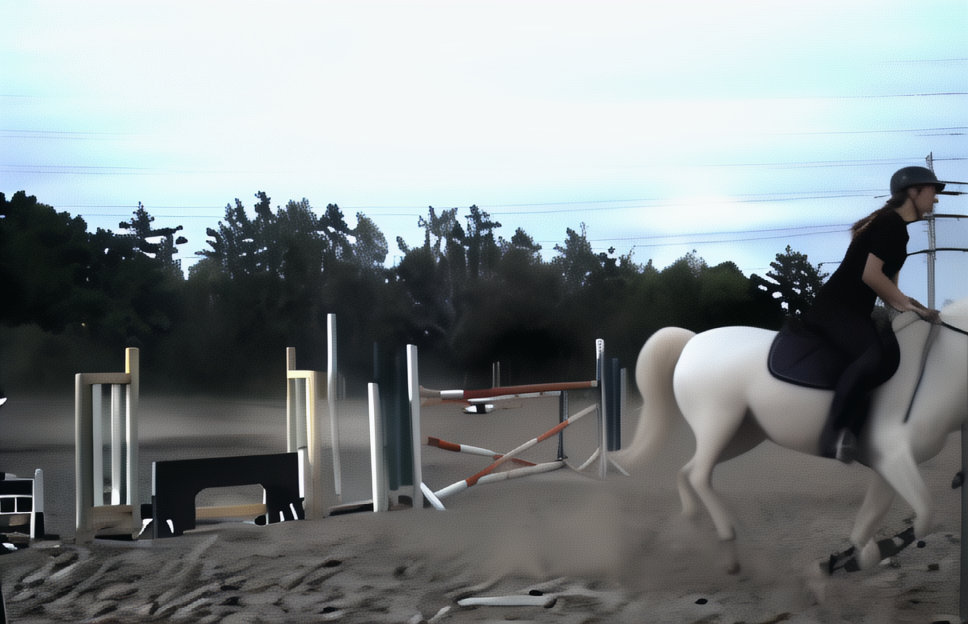}} & \tile{\includegraphics[width=0.161\textwidth]{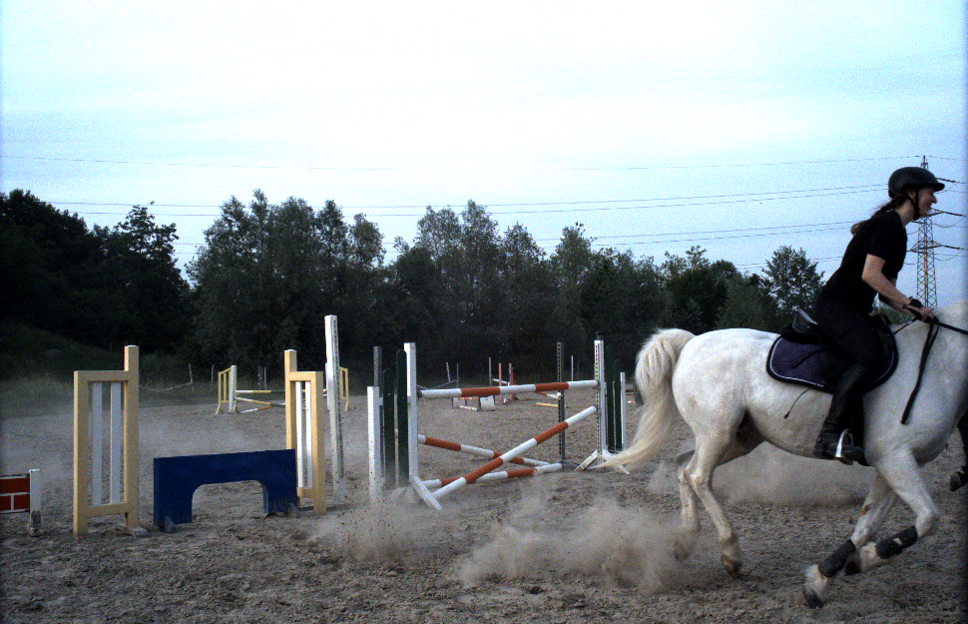}}\\[3pt]
\lab{E2VID+} & \lab{EvDiff} & \lab{EDI} & \lab{UniE2F$^\dagger$} & \lab{\textbf{Ours}} & \lab{GT} \\
\end{tabular}
\vspace{2pt}   \caption{\textbf{Qualitative comparison on BS-ERGB.} Frames 74 and 77 of each method. \ourmethod reconstructs consistent brightness and colour, while events-only baselines exhibit saturation or low contrast; given the same map, EDI stays close to the static map and misses the galloping horse, and UniE2F$^\dagger$ blurs the horse and ghosts the rider. The supplementary material shows all compared methods.}
\label{fig:qual_bsergb_horse}
\end{figure*}

Existing datasets~\cite{E2VID+,zhu2018multivehicle,tulyakov2021time,tulyakov2022time} record motion events without temporal-mapping events, making them unsuitable for our dual-stream reconstruction paradigm.
To realise this paradigm, we implement a hardware setup (Fig.~\ref{fig:dataset}; components in the supplementary material) and collect \ourdataset, which pairs temporal-mapping and motion events with synchronised RGB ground truth.

The event and RGB sensors are synchronised by hardware triggering. In contrast to prior dual-camera designs~\cite{tulyakov2021time} and beam-splitter--first configurations~\cite{tulyakov2022time}, our 50R/50T beam splitter sits between the lens, which carries the motorised aperture, and the sensors. Both sensors therefore share one centre of projection, so a single homography~\cite{hartley2003multiple} aligns them independently of scene depth.

\ourdataset consists of $99$ sequences with $14{,}307$ RGB images and the corresponding events at $1280\times720$. It covers indoor and outdoor scenes with object and camera motion and non-rigid dynamics such as liquids, whose highly deformable structure is challenging for video reconstruction.

\section{Experiments}
\label{sec:experiments}
\begin{table*}[t]
\centering
\small
\setlength{\tabcolsep}{4pt}
\renewcommand\arraystretch{1.0}   \caption{\textbf{Quantitative comparison on \ourdataset.} An initial temporal map recovered from genuine events improves reconstruction fidelity and perceptual quality. CogVideoX-2B$^{*}$ and UniE2F$^\dagger$ are defined in Tab.~\ref{tab:bsergb}.}
\begin{tabular}{r||ccccc?ccccc}
    \thickhline
    \rowcolor{tableHeadGray}
    & \multicolumn{5}{c?}{\textbf{RGB}} & \multicolumn{5}{c}{\textbf{Luminance ($Y$)}} \\
    \rowcolor{tableHeadGray}
    \multirow{-2}{*}{\textbf{Method}}
    & \textbf{PSNR $\uparrow$} & \textbf{SSIM $\uparrow$} & \textbf{LPIPS $\downarrow$} & \textbf{FID $\downarrow$} & \textbf{FVD $\downarrow$}
    & \textbf{PSNR $\uparrow$} & \textbf{SSIM $\uparrow$} & \textbf{LPIPS $\downarrow$} & \textbf{FID $\downarrow$} & \textbf{FVD $\downarrow$} \\
    \hline
    \rowcolor{tableSubHeadGray}
    SPADE-E2VID\pub{TIP2021}~\cite{Cadena2021Spade-E2VID} & 8.26 & 0.2248 & 0.8088 & 346.2 & 2631.0 & 8.30 & 0.2487 & 0.7453 & 307.0 & 2000.0 \\
    ET-Net\pub{ICCV2021}~\cite{ET-Net}                   & 15.80 & 0.4911 & 0.6539 & 365.6 & 2221.2 & 16.00 & 0.5411 & 0.5612 & 325.9 & 1606.6 \\
    \rowcolor{tableSubHeadGray}
    E2VID+\pub{ECCV2020}~\cite{E2VID+}                   & 17.79 & 0.5657 & 0.5274 & 340.7 & 2248.6 & 18.12 & 0.6204 & 0.4454 & 307.1 & 1604.9 \\
    FireNet\pub{WACV2020}~\cite{2020FireNet}             & 7.21 & 0.2739 & 0.8020 & 369.3 & 3039.4 & 7.23 & 0.3095 & 0.7418 & 331.0 & 2379.3 \\
    \rowcolor{tableSubHeadGray}
    EvDiff\pub{ECCV2026}~\cite{Li2025EvDiffEV}           & 17.70 & 0.5342 & \fs 0.4011 & 350.2 & 2555.4 & 18.00 & 0.6091 & 0.4068 & 326.9 & 2068.2 \\
    UniE2F\pub{arXiv2026}~\cite{unie2f}                  & 12.75 & 0.4485 & 0.6331 & 322.9 & 2453.2 & 12.87 & 0.4884 & 0.5654 & 307.0 & 2064.8 \\
    \rowcolor{tableSubHeadGray}
    LongE2V\pub{SIGGRAPH2026}~\cite{fan2026longe2v}      & 11.35 & 0.3559 & 0.7125 & 322.8 & 2436.6 & 11.42 & 0.3993 & 0.6523 & 295.9 & 1917.3 \\
    CogVideoX-2B$^{*}$\pub{ICLR2025}~\cite{yang2024cogvideox}          & 15.34 & 0.5660 & 0.5133 & 266.4 & 1756.2 & 15.51 & 0.6224 & 0.4781 & 257.4 & 1413.5 \\
    \hline\hline
    \multicolumn{11}{l}{\textcolor{gray}{\textit{Given the same initial temporal map as \ourmethod}}} \\
    \rowcolor{tableSubHeadGray}
    \quad EDI\pub{CVPR2019}~\cite{edi_pan}                             & 13.09 & 0.1461 & 0.9964 & 282.7 & 2726.0 & 13.19 & 0.1529 & 0.9625 & 247.5 & 2124.2 \\
    \quad UniE2F$^\dagger$\pub{arXiv2026}~\cite{unie2f}  & \nd 18.63 & \nd 0.6435 & 0.4459 & \nd 228.7 & \nd 1716.8 & \nd 18.99 & \nd 0.7034 & \nd 0.3701 & \nd 221.4 & \nd 1369.3 \\
    \hline\hline
    \textbf{\ourmethod~(Ours)}                           & \fs 19.88 & \fs 0.6654 & \nd 0.4368 & \fs 183.3 & \fs 1575.8 & \fs 20.55 & \fs 0.7625 & \fs 0.3387 & \fs 171.1 & \fs 1272.7 \\
    \thickhline
\end{tabular}
\label{tab:ourdataset}
\vspace{-1pt}   \end{table*}

\begin{figure*}[t]
\centering
\setlength{\tabcolsep}{0.9pt}
\renewcommand{\arraystretch}{0.1}

\setlength{\fboxsep}{0pt}\setlength{\fboxrule}{0.3pt}
\newcommand{\tile}[1]{\fcolorbox{black!30}{white}{#1}}
\newcommand{\lab}[1]{\multicolumn{1}{c}{\footnotesize #1}}

\begin{tabular}{cccccc}
\tile{\includegraphics[width=0.161\textwidth]{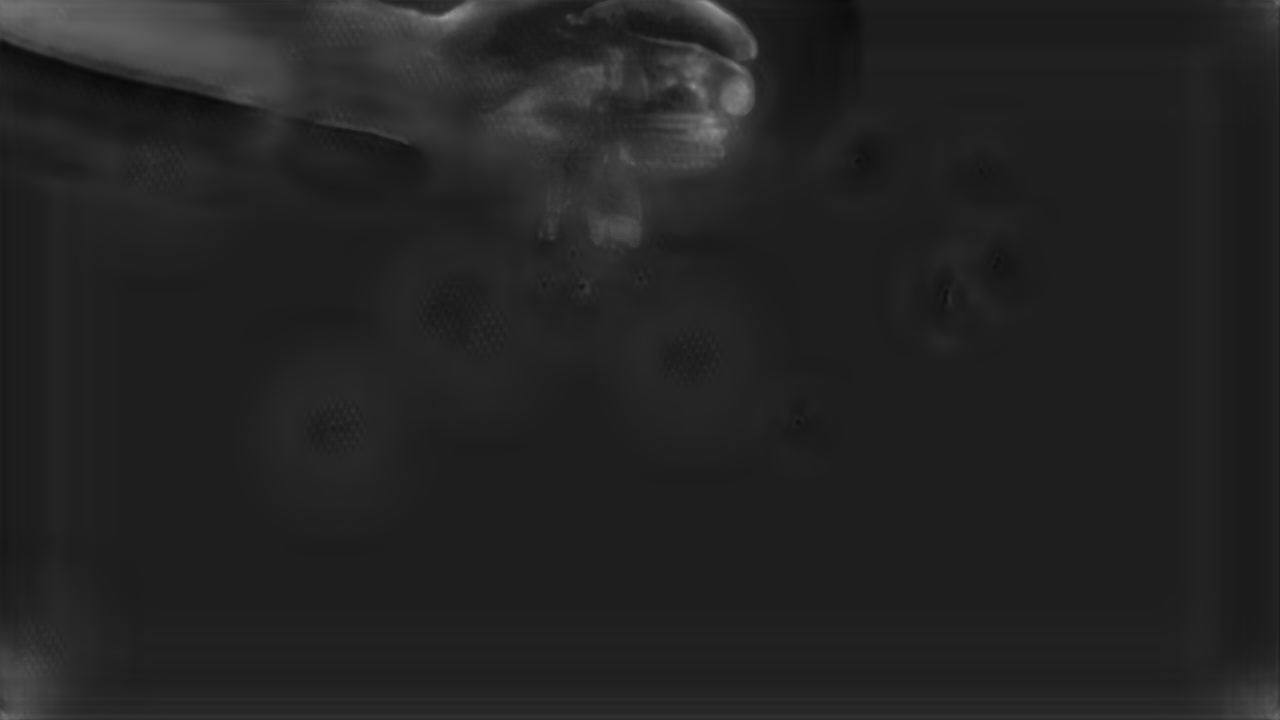}} & \tile{\includegraphics[width=0.161\textwidth]{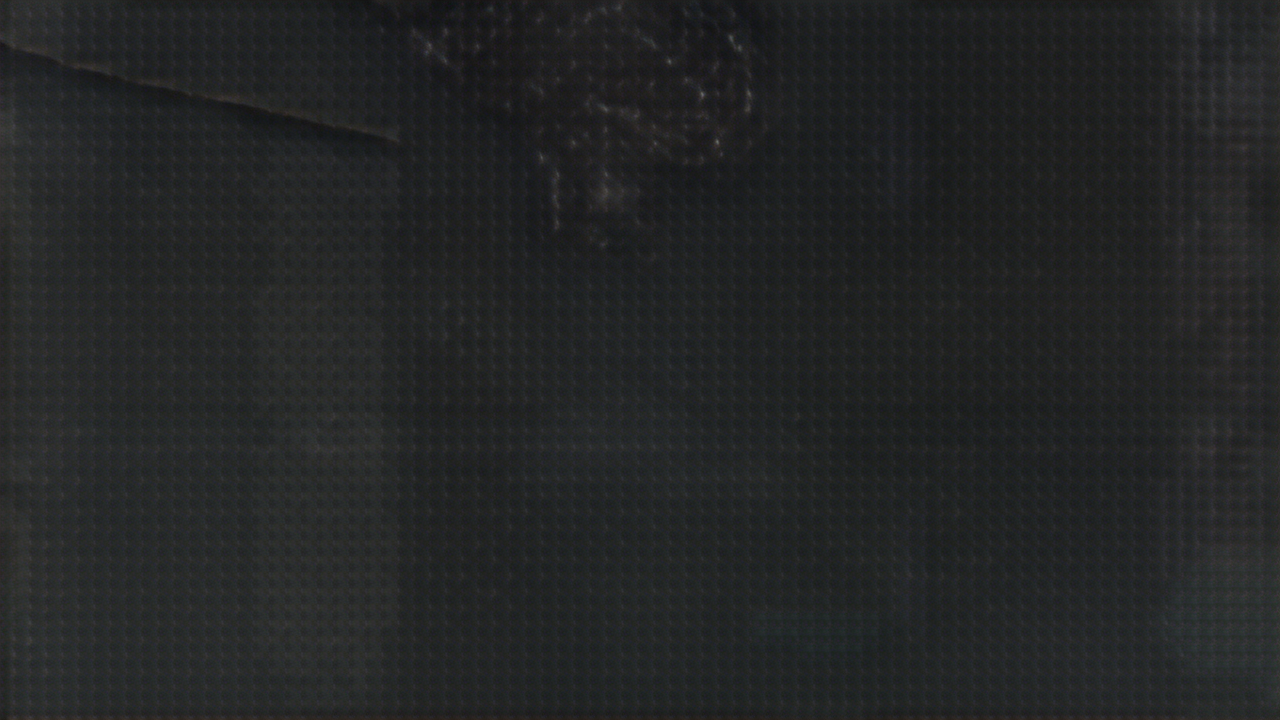}} & \tile{\includegraphics[width=0.161\textwidth]{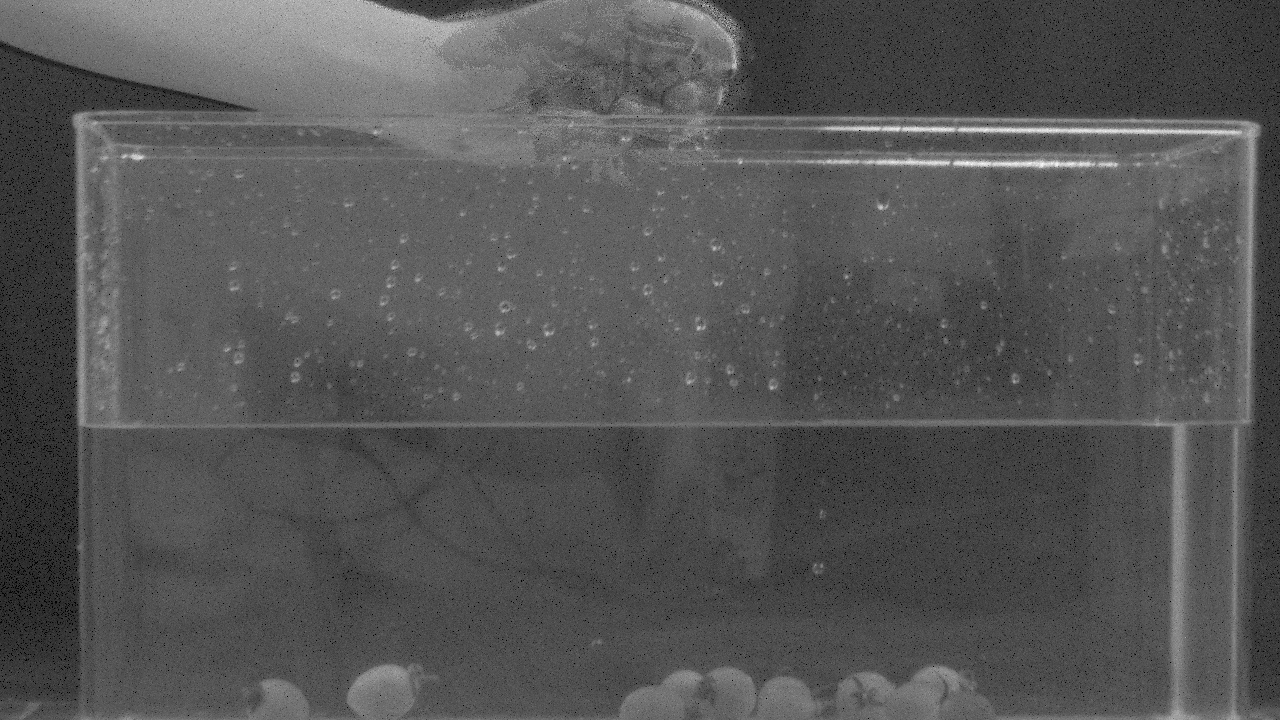}} & \tile{\includegraphics[width=0.161\textwidth]{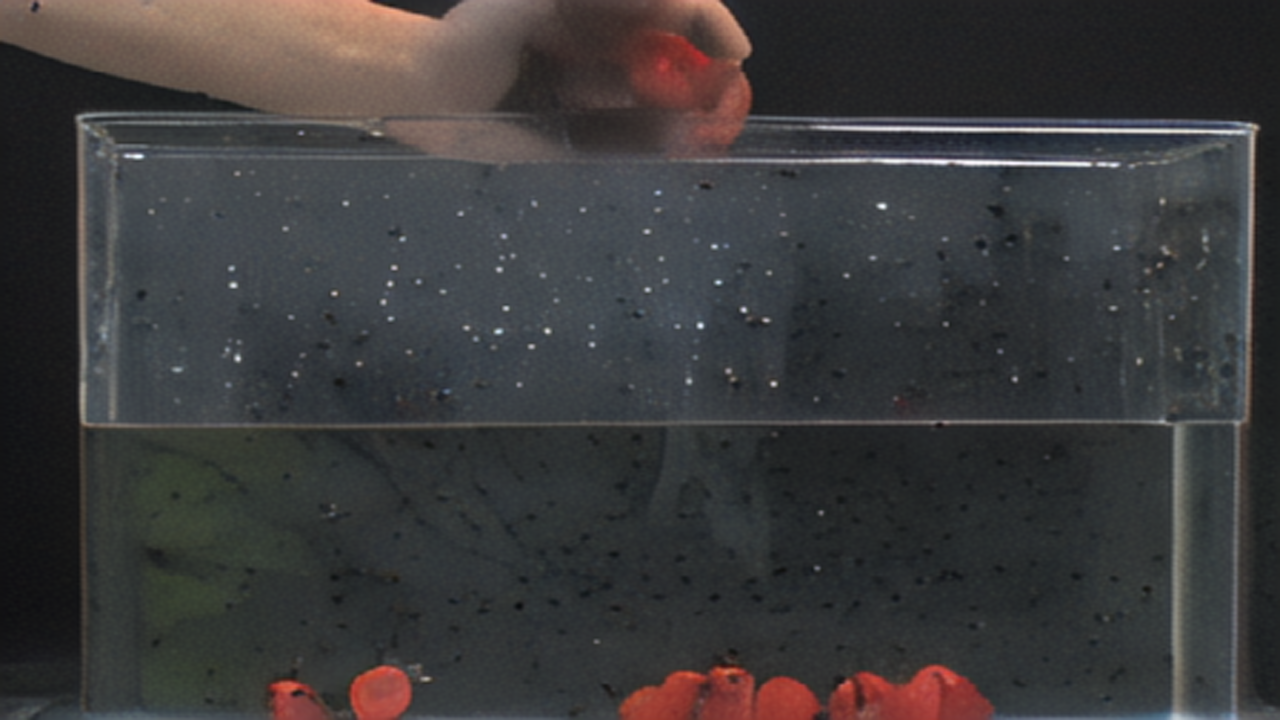}} & \tile{\includegraphics[width=0.161\textwidth]{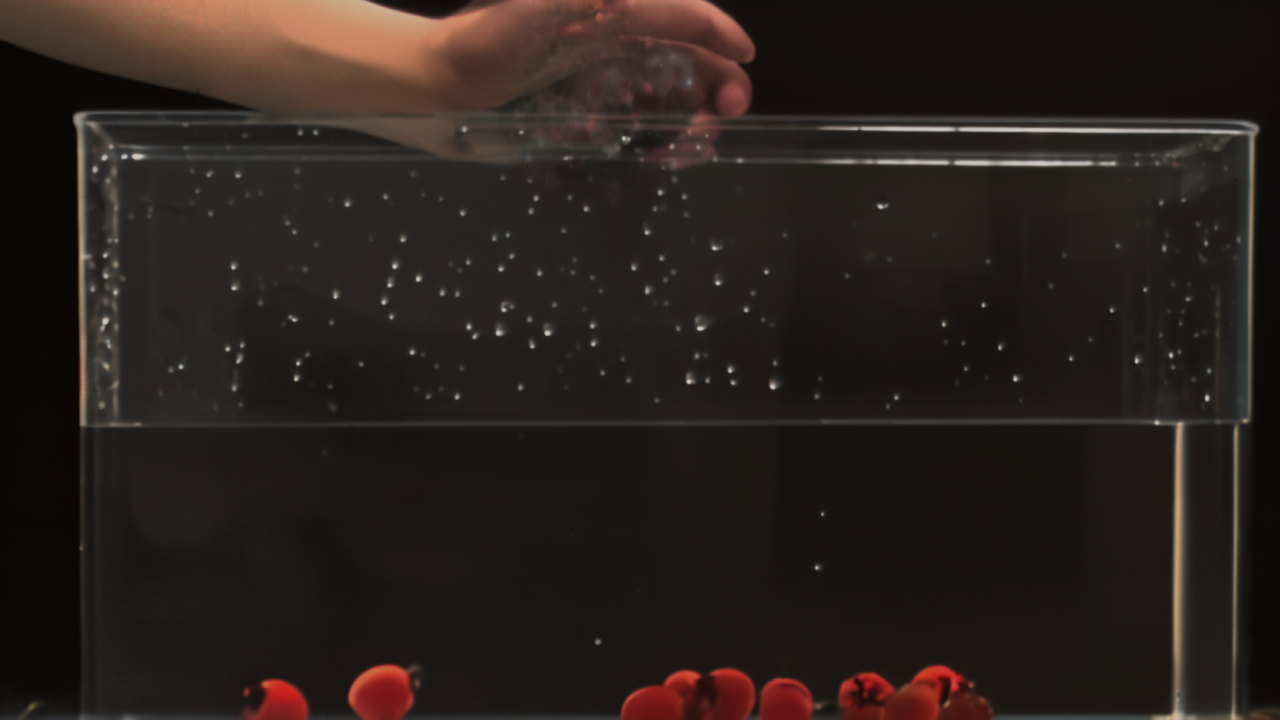}} & \tile{\includegraphics[width=0.161\textwidth]{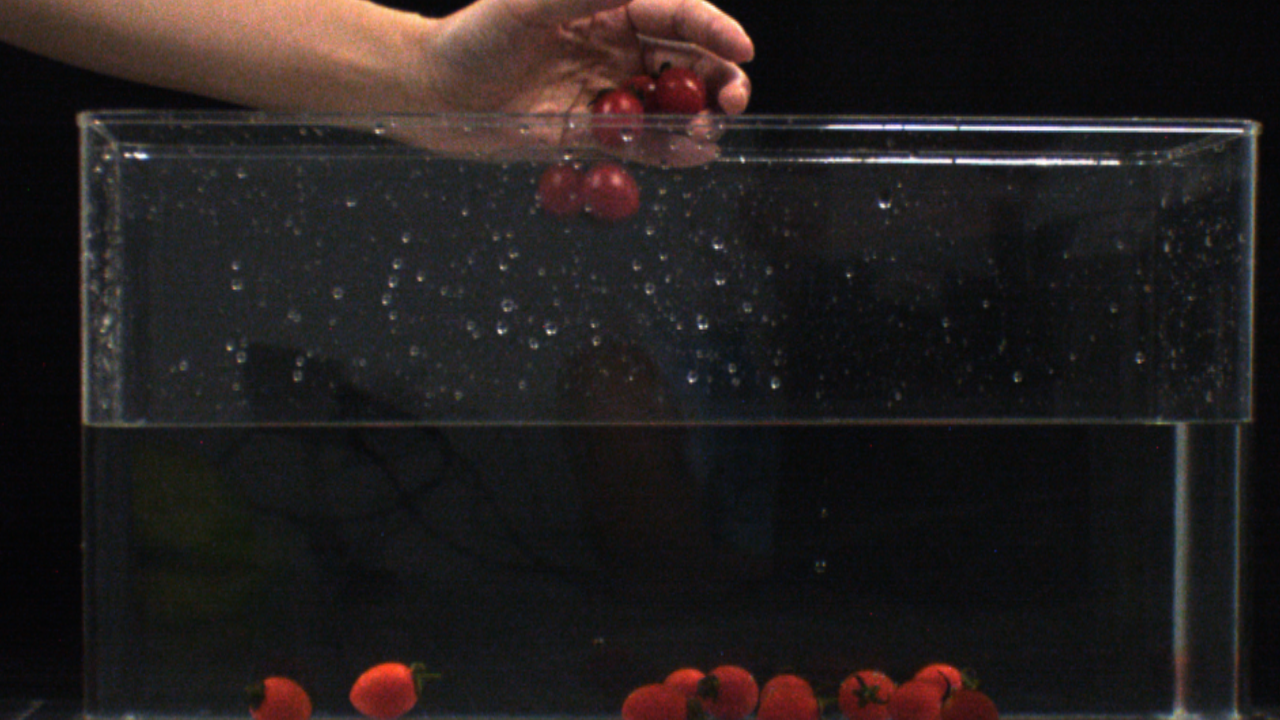}}\\[1.5pt]
\tile{\includegraphics[width=0.161\textwidth]{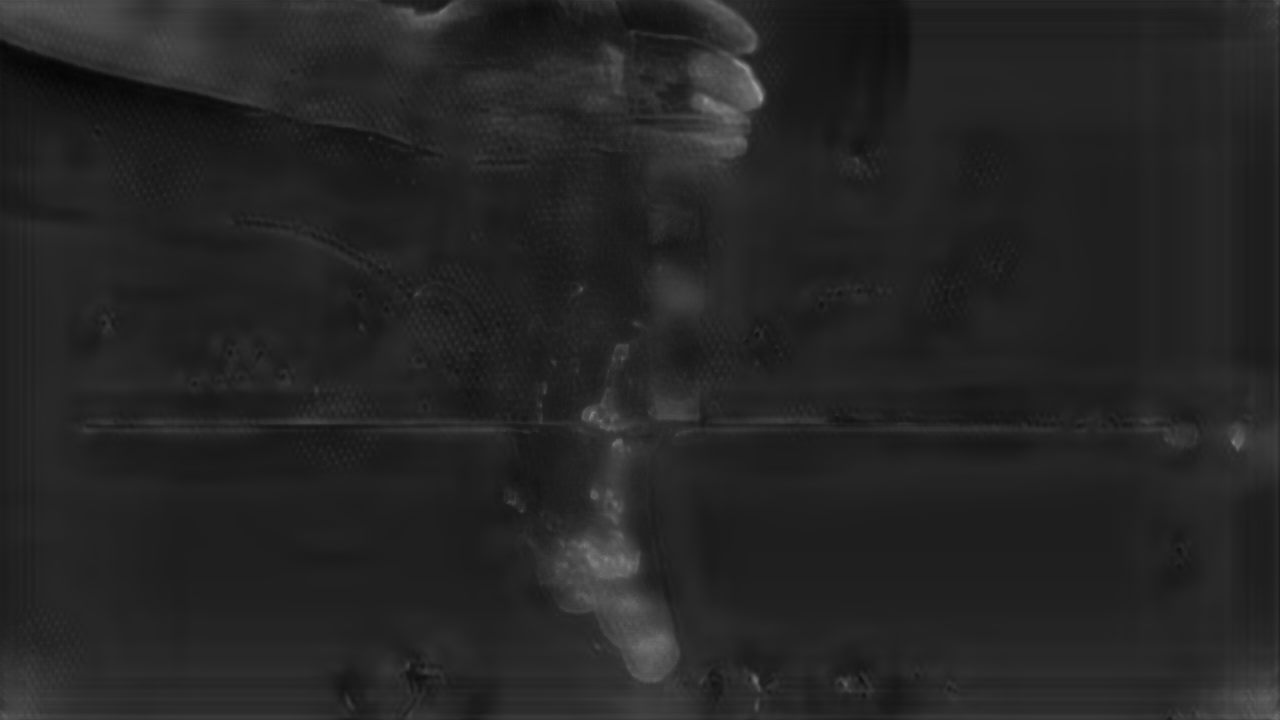}} & \tile{\includegraphics[width=0.161\textwidth]{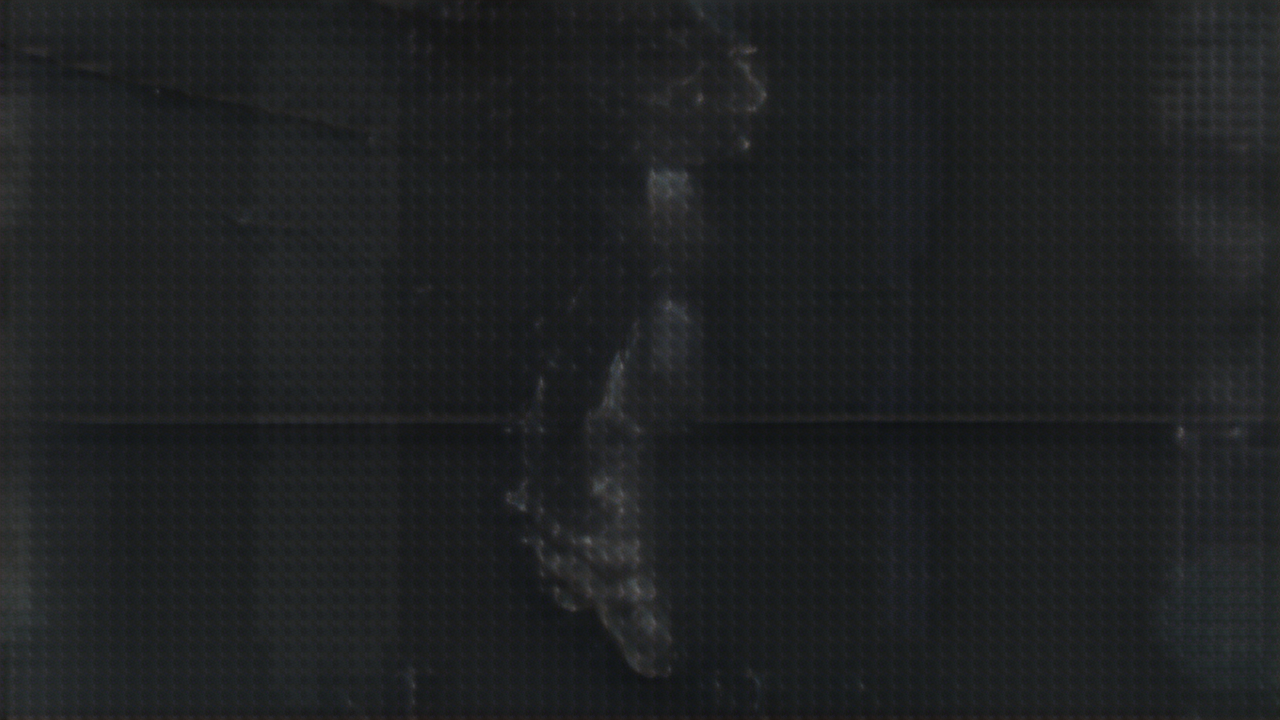}} & \tile{\includegraphics[width=0.161\textwidth]{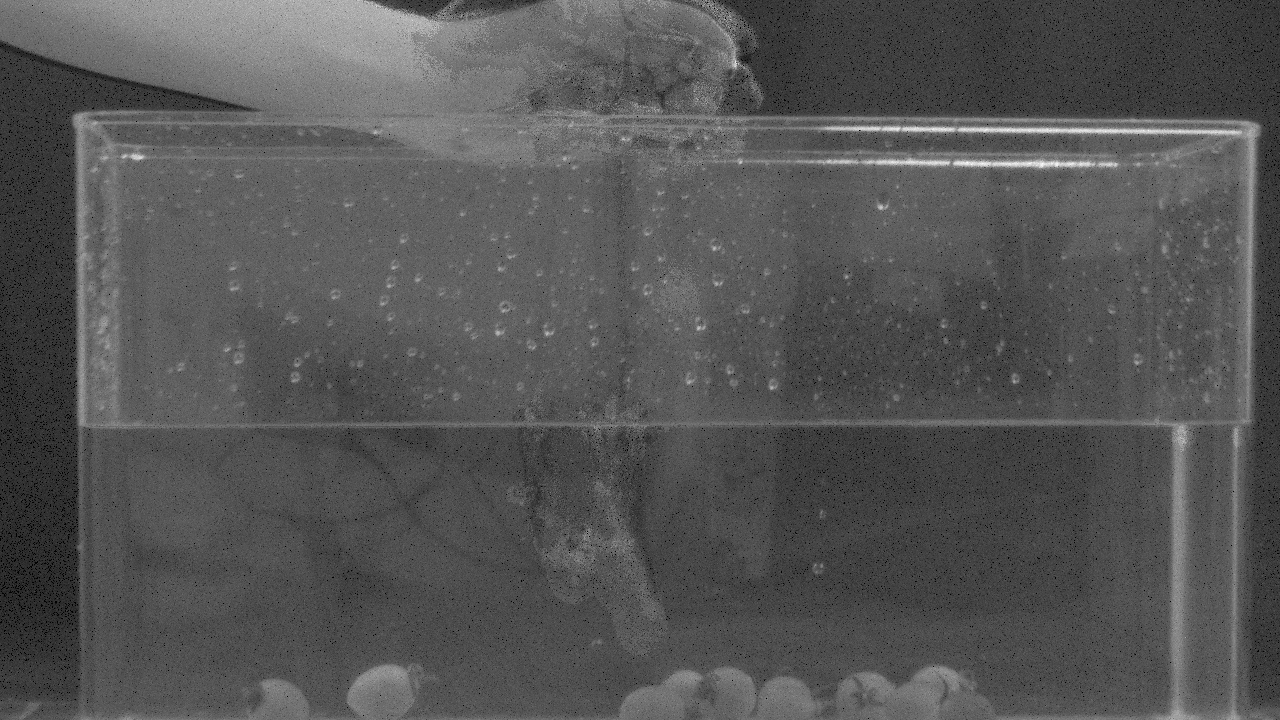}} & \tile{\includegraphics[width=0.161\textwidth]{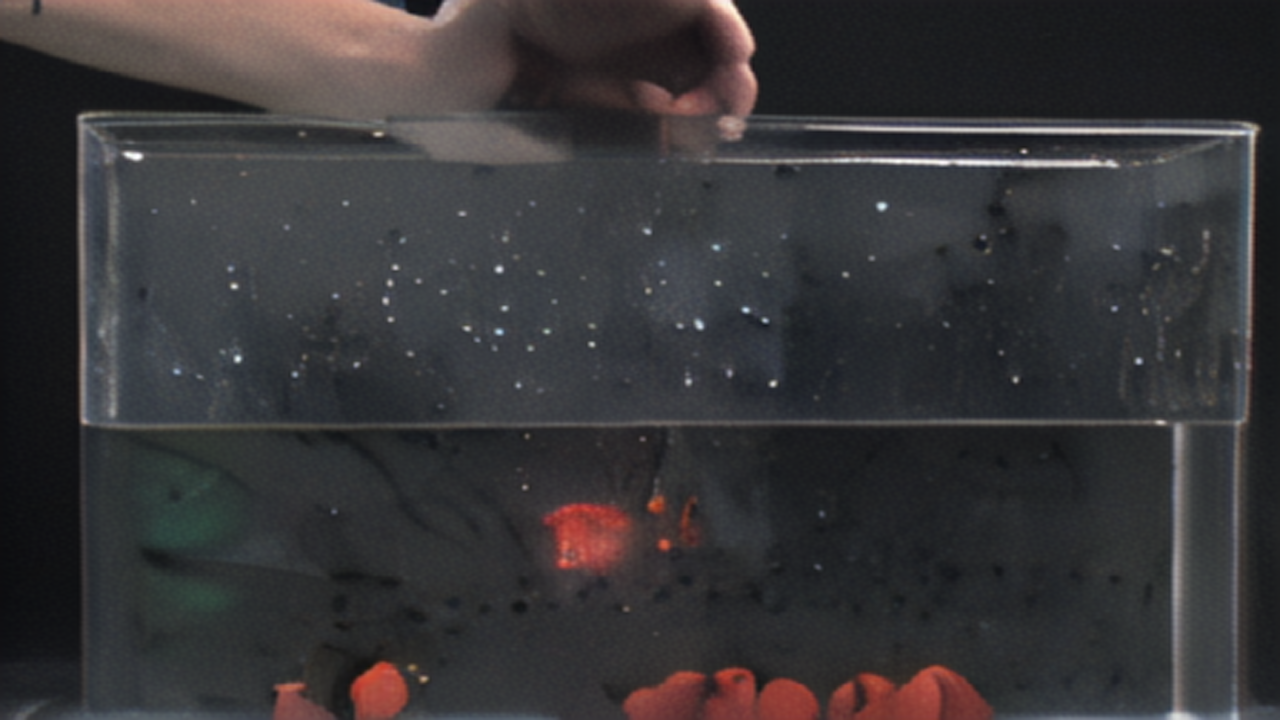}} & \tile{\includegraphics[width=0.161\textwidth]{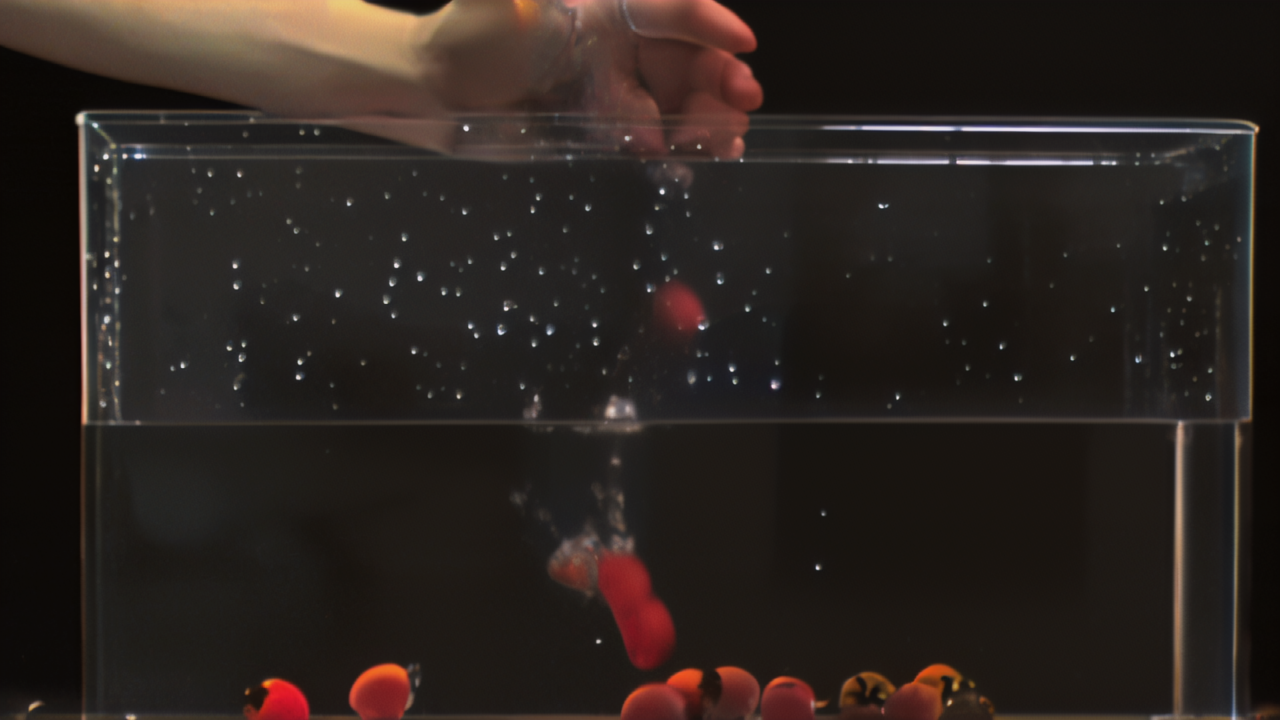}} & \tile{\includegraphics[width=0.161\textwidth]{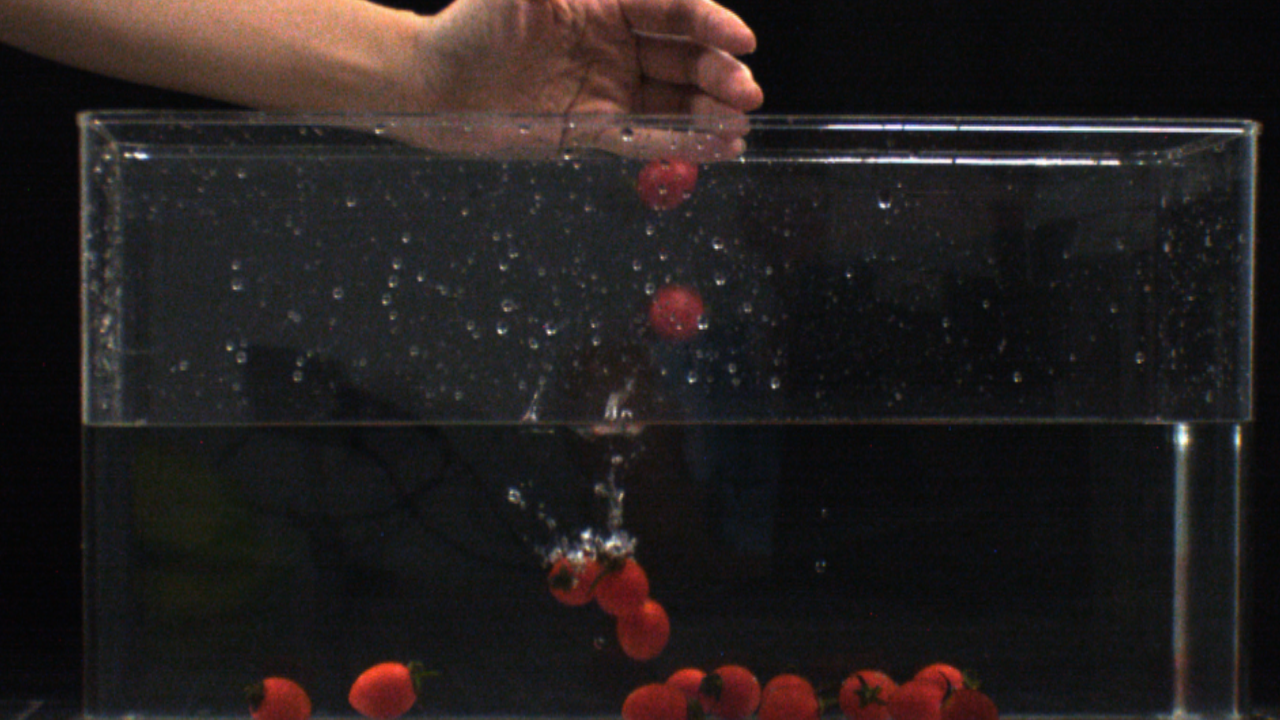}}\\[3pt]
\lab{E2VID+} & \lab{EvDiff} & \lab{EDI} & \lab{UniE2F$^\dagger$} & \lab{\textbf{Ours}} & \lab{GT} \\
\end{tabular}
\vspace{2pt}   \caption{\textbf{Qualitative comparison on \ourdataset.} Frames 8 (first chunk) and 14 (second chunk) of each method. Reconstruction from one temporal map preserves scene appearance across successive video chunks; given the same map, EDI stays close to the static map and UniE2F$^\dagger$ merges the falling tomatoes into one blob. The supplementary material shows all compared methods.}
\label{fig:qual_dvs_seq12}
\end{figure*}

\subsection{Datasets and Evaluation Metrics}

\PAR{BS-ERGB Dataset.} It provides RGB sequences with a spatial resolution of $970 \times 625$ and contains a total of $123$ sequences. We use $46$ sequences for training and $26$ for testing, with $30$ to $531$ frames each and paired events.
As BS-ERGB contains no temporal-mapping events, we follow~\cite{sun2025low} and synthesise the initial map of each sequence by applying the degradation model $D$ of Sec.~\ref{sec:degradation} to the luminance of its first frame; the map is single-channel, and the two map-conditioned baselines receive the same map.

\PAR{\ourdataset Dataset.} It provides RGB sequences with a spatial resolution of $1280 \times 720$ and contains $99$ sequences (Sec.~\ref{sec:dataset}), with $6$ to $406$ frames each and measured temporal-mapping and motion events. We use $75$ sequences/$10{,}552$ frames for training and $24$ sequences/$3{,}755$ frames for validation. Six training sequences are shorter than one chunk, so the control branch is trained on the other $69$ ($10{,}497$ frames).

\PAR{Evaluation Protocol.}
\label{sec:protocol}
We use PSNR and SSIM~\cite{wang2004image} for pixel fidelity, LPIPS~\cite{zhang2018unreasonable} and FID~\cite{heusel2017gans} for perceptual quality, and FVD~\cite{unterthiner2018towards} for video-level realism.
Our method, EvDiff, UniE2F and the CogVideoX-2B baseline output colour videos, whereas the other baselines output greyscale, so we report both RGB and luminance metrics; the conversions are specified in the supplementary material.
Under the \emph{single-map protocol}, each sequence receives one temporal map at its start and every later chunk a self-made map.

\PAR{Compared Methods.}~The events-only baselines are E2VID+~\cite{E2VID+}, FireNet~\cite{2020FireNet}, SPADE-E2VID~\cite{Cadena2021Spade-E2VID}, ET-Net~\cite{ET-Net}, EvDiff~\cite{Li2025EvDiffEV}, UniE2F~\cite{unie2f} and LongE2V~\cite{fan2026longe2v}.
We follow the training settings recommended in the respective papers and finetune public weights on each training split. UniE2F and LongE2V use released weights on BS-ERGB. Both run in events-only mode.
To test whether a stronger video prior closes the gap, we train an event ControlNet on a frozen CogVideoX-2B~\cite{yang2024cogvideox} text-to-video model for each dataset. This baseline, CogVideoX-2B$^{*}$, reconstructs $13$-frame chunks without a reference frame.
Two baselines receive the same temporal map as \ourmethod: EDI~\cite{edi_pan}, which integrates events onto the log intensity of the map without the deblurring integral, with thresholds fitted on each training split, and UniE2F$^\dagger$, which is retrained with the map as its reference frame and re-anchored on its own output in every later chunk, like \ourmethod.
Baseline training budgets and settings are detailed in the supplementary material.
We exclude E2VIDiff~\cite{Liang2024E2VIDiffPE}, as neither its source code nor trained models are publicly available.

\subsection{Implementation Details}
We implement \ourmethod with a frozen SVD backbone and train per dataset on one NVIDIA H200 GPU, using AdamW with a learning rate of $5\times10^{-6}$ on $512\times320$ crops.
The \ourencoder is initialised from the SVD VAE encoder and trained by latent alignment.
We then train the control branch on $13$-frame chunks with an effective batch size of $128$ using FP16.
Finally, we finetune this branch using maps generated from its own training-sequence rollouts.
More details kindly refer to \textit{supp}.

\subsection{Comparisons on the BS-ERGB Dataset}
\begin{figure}[t]
\centering
\includegraphics[width=\columnwidth]{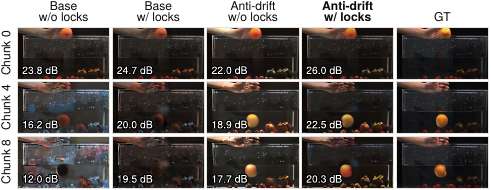}
\caption{\textbf{Effects of anti-drift finetuning and photometric locks.} Finetuning preserves content and the locks stabilise brightness; our final setting is in bold. Labels give luminance PSNR.}
\label{fig:streaming}
\end{figure}

\begin{figure*}[t]
\centering
\includegraphics[width=\textwidth]{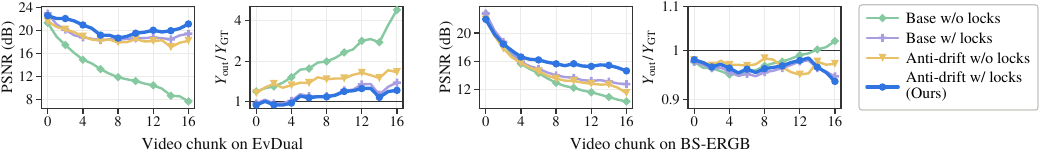}
\caption{\textbf{Drift along the rollout.} Luminance PSNR and mean luminance ratio $Y_{\mathrm{out}}/Y_{\mathrm{GT}}$ across video chunks, averaged over sequences reaching each chunk. The ratio uses a logarithmic scale. Photometric locks limit brightness drift, and our method achieves the highest PSNR in later chunks on both datasets.}
\label{fig:drift}
\end{figure*}

Tab.~\ref{tab:bsergb} shows that \ourmethod leads on every RGB and luminance metric, with luminance PSNR of $17.35$ versus $16.54$~dB, SSIM of $0.50$ versus $0.45$, and FVD of $954$ versus $1446$ against the best baseline for each metric.
Among the events-only baselines, CogVideoX-2B$^{*}$ leads on no metric here: LongE2V is ahead in SSIM, FID and FVD, and EvDiff in PSNR and LPIPS.
With the same map, UniE2F$^\dagger$ reaches $16.33$~dB, up from $10.16$~dB for UniE2F, and EDI reaches $16.54$~dB while staying close to the static map, yet both trail in SSIM and FVD, so the map alone does not account for our margin.

Fig.~\ref{fig:qual_bsergb_horse} shows that the events-only baselines E2VID+ and EvDiff recover the moving horse but neither the static background nor the brightness of the scene.
Our reconstruction retains stable colour $74$ frames after the temporal map, illustrating the benefit of a single brightness reference.

\subsection{Comparisons on the \ourdataset Dataset}

In Tab.~\ref{tab:ourdataset}, \ourmethod leads on every luminance metric, with PSNR of $20.55$ versus $18.99$~dB, SSIM of $0.76$ versus $0.70$, and FVD of $1273$ versus $1369$ against the best baseline for each metric, and on every RGB metric except LPIPS, where EvDiff scores $0.40$ against our $0.44$.
Among the events-only baselines, CogVideoX-2B$^{*}$ ranks first in SSIM, FID and FVD, yet reaches only $15.51$~dB luminance PSNR, below E2VID+, EvDiff and ET-Net: a stronger video prior alone improves realism but not fidelity.
With the same temporal map, UniE2F$^\dagger$ reaches $18.99$~dB, up from $12.87$~dB for UniE2F, but trails on every metric; EDI reaches $13.19$~dB.
Fig.~\ref{fig:qual_dvs_seq12} illustrates reconstruction from a single temporal map, with further examples provided in the supplementary material.

\subsection{Ablation Study}
\label{sec:ablation}
\begin{table}[t]
\centering
\small
\setlength{\tabcolsep}{4pt}
\renewcommand\arraystretch{1.12}   \caption{\textbf{Base-training ablation on \ourdataset.} Variants share the training schedule of \emph{Base} before anti-drift finetuning. The temporal-map ablation omits the map during training and evaluation and disables both locks.}
\resizebox{\columnwidth}{!}{    \begin{tabular}{l||ccccc}
    \thickhline
    \rowcolor{tableHeadGray}
    \textbf{Method}
    & \textbf{PSNR $\uparrow$}
    & \textbf{SSIM $\uparrow$}
    & \textbf{LPIPS $\downarrow$}
    & \textbf{FID $\downarrow$}
    & \textbf{FVD $\downarrow$} \\
    \hline
    \rowcolor{tableSubHeadGray}
    \quad \textit{w/o} encoder tuning    & 17.44 & 0.5832 & 0.4977 & 285.6 & 1876.7 \\
    \quad \textit{w/o} degradation       & \nd 19.27 & 0.6832 & \nd 0.4185 & 220.3 & 1345.9 \\
    \rowcolor{tableSubHeadGray}
    \quad \textit{w/o} both locks        & 15.35 & 0.5158 & 0.4915 & \fs 214.4 & 1504.5 \\
    \quad \textit{w/o} temporal map      & 17.91 & \nd 0.7090 & 0.4226 & \nd 215.1 & \fs 1188.0 \\
    \hline\hline
    \textbf{Base}                        & \fs 19.69 & \fs 0.7190 & \fs 0.4073 & 226.5 & \nd 1289.4 \\
    \thickhline
    \end{tabular}
}
\label{tab:ablation_base}
\vspace{-6pt}
\end{table}

\begin{table}[t]
\centering
\small
\setlength{\tabcolsep}{4pt}
\renewcommand\arraystretch{1.03}   \caption{\textbf{Ablation of the final model on \ourdataset, luminance.} Each row changes one setting of the final model, the last row.}
\resizebox{\columnwidth}{!}{    \begin{tabular}{l||ccccc}
    \thickhline
    \rowcolor{tableHeadGray}
    \textbf{Method}
    & \textbf{PSNR $\uparrow$}
    & \textbf{SSIM $\uparrow$}
    & \textbf{LPIPS $\downarrow$}
    & \textbf{FID $\downarrow$}
    & \textbf{FVD $\downarrow$} \\
    \hline
    \multicolumn{6}{l}{\textcolor{gray}{\textit{Conditioning}}} \\
    \rowcolor{tableSubHeadGray}
    \quad \textit{w/o} motion events     & 16.54 & 0.7056 & 0.4271 & 240.4 & 2088.1 \\
    \quad \textit{w/o} CLIP              & 18.82 & 0.7430 & 0.3552 & 185.2 & 1487.8 \\
    \hline
    \multicolumn{6}{l}{\textcolor{gray}{\textit{Tiled inference}}} \\
    \rowcolor{tableSubHeadGray}
    \quad Single view (no tiling)        & 19.11 & 0.6962 & 0.4012 & 190.3 & 1434.9 \\
    \quad \textit{w/o} weight floor      & 18.74 & 0.7370 & 0.3695 & 182.3 & 1334.2 \\
    \rowcolor{tableSubHeadGray}
    \quad Uniform weights                & 20.23 & \fs 0.7628 & 0.3595 & 185.9 & 1351.2 \\
    \hline
    \multicolumn{6}{l}{\textcolor{gray}{\textit{Photometric locks}}} \\
    \rowcolor{tableSubHeadGray}
    \quad \textit{w/o} colour lock       & 19.50 & 0.7576 & 0.3436 & 175.3 & 1451.8 \\
    \quad \textit{w/o} low-band freeze   & \nd 20.35 & 0.7567 & \nd 0.3418 & 174.5 & \nd 1197.6 \\
    \rowcolor{tableSubHeadGray}
    \quad \textit{w/o} both locks        & 19.23 & 0.7259 & 0.3539 & 178.6 & 1391.2 \\
    \hline
    \multicolumn{6}{l}{\textcolor{gray}{\textit{Anti-drift finetuning}}} \\
    \rowcolor{tableSubHeadGray}
    \quad $25\%$ self-made maps          & 20.15 & 0.7509 & 0.3534 & \fs 170.3 & \fs 1190.3 \\
    \quad Clean maps, same steps         & 19.59 & 0.7125 & 0.3985 & 221.9 & 1307.7 \\
    \rowcolor{tableSubHeadGray}
    \quad \textit{w/o} finetuning (Base) & 19.69 & 0.7190 & 0.4073 & 226.5 & 1289.4 \\
    \hline\hline
    \textbf{\ourmethod~(Final)}          & \fs 20.55 & \nd 0.7625 & \fs 0.3387 & \nd 171.1 & 1272.7 \\
    \thickhline
    \end{tabular}
}
\label{tab:ablation}
\vspace{-6pt}
\end{table}

\PAR{Base Training.}~Relative to \emph{Base} (Tab.~\ref{tab:ablation_base}), replacing the Time-to-Brightness Encoder with the original VAE encoder costs $2.3$~dB, and the latent mismatch propagates through self-made maps.
Training on plain greyscale maps without degradation costs $0.4$~dB.
The locks improve the base model by $4.3$~dB versus $1.3$~dB after finetuning, indicating that finetuning already reduces photometric drift.
Under the same training schedule, nulling the temporal-map latent and CLIP embedding in every sample and chunk, without locks, yields $17.91$~dB, $1.8$~dB below \emph{Base}.

\PAR{Conditioning and Fusion.}~Relative to the final model (Tab.~\ref{tab:ablation}), removing motion events reduces PSNR by $4.0$~dB and worsens all other metrics: a temporal map supplies appearance but cannot describe scene evolution.
Removing CLIP image conditioning costs $1.7$~dB, showing that semantic features complement event inputs.
Single-view denoising, removing the weight floor and uniform weights reduce PSNR by $1.4$, $1.8$ and $0.3$~dB, respectively.

\PAR{Photometric Locks.}~Disabling both locks costs $1.3$~dB. Removing only the colour lock, including its chunk-$0$ gain, costs $1.05$~dB, versus $0.2$~dB for removing only the low-band freeze.
Together they give the best PSNR and LPIPS, although individual metrics favour different variants.
Fig.~\ref{fig:streaming} shows that finetuning preserves content while the locks correct brightness bias.
Across all sequences (Fig.~\ref{fig:drift}), the base output without locks is $4.81$ times as bright as ground truth. Finetuning alone limits this ratio to $1.71$, while the locks keep it within $0.95$--$1.26$.

\PAR{Anti-Drift Finetuning.}~Without finetuning (\emph{Base}), PSNR drops by $0.9$~dB, SSIM falls from $0.76$ to $0.72$, LPIPS rises from $0.34$ to $0.41$, and FID from $171$ to $227$.
Extra base training with clean-derived maps for the same $3{,}150$ steps yields $19.59$~dB, SSIM $0.71$ and LPIPS $0.40$, on par with \emph{Base}. Additional training alone does not explain the gain.
Replacing only $25\%$ of training maps with self-made maps, rather than all, yields $20.15$~dB and the best FID and FVD, but lower SSIM and higher LPIPS than the final model.

\section{Conclusion}
We presented \ourmethod, a dual-stream framework inspired by complementary processing in the M and P pathways. It conditions video diffusion on temporal-mapping and motion events for autoregressive reconstruction from one temporal map. Photometric constraints and rollout finetuning reduce accumulated drift. We also built a capture system and collected \ourdataset\ with synchronised dual-stream events and ground-truth RGB videos. Experiments on \ourdataset\ and BS-ERGB demonstrate improved fidelity, perceptual quality and video-level realism.

\PAR{Limitations.}~Content drift persists as errors propagate through self-made maps (Fig.~\ref{fig:drift}). Acquisition requires a modulation device and discards mixed events during transitions, preventing direct use of unmodulated event streams.

\pagebreak
\section*{LLM Usage Statement}
We used Anthropic Claude (Fable and Opus models) for assistance with writing, coding, and experiments.
We used Google Deep Review for pre-submission feedback on claims and clarity.
The authors designed the study, conducted the experiments, verified all reported results, and take full responsibility for the paper.

\end{document}